\PassOptionsToPackage{hyperfootnotes=false}{hyperref}
\documentclass[a4paper,fleqn]{cas-sc}
\usepackage{mathtools}
\usepackage[authoryear,longnamesfirst]{natbib}
\let\printorcid\relax
\hypersetup{hidelinks}
\makeatletter
\let\ps@first\ps@plain
\makeatother

\newcommand{\meanstd}[2]{%
  \ensuremath{#1\,{\text{\fontsize{7pt}{7pt}\selectfont(\textpm\,#2)}}}%
}
\newcommand{\papertitle}{%
  Reinforcement learning for inverse silhouette design and rapid laser cutting of compact parallelogram quad kirigami prototypes%
}

\ExplSyntaxOn
\NewDocumentCommand{\casnologo}{}{\keys_set:nn { stm / mktitle } { nologo }}
\ExplSyntaxOff

\begin{document}
\let\WriteBookmarks\relax
\def\floatpagepagefraction{1}
\def\textpagefraction{.001}
\pagestyle{plain}

\shorttitle{}
\shortauthors{M. Yazdani et al.}
\title[mode=title]{\papertitle\tnotemark[1]}

\author[aff1,aff4]{Milad Yazdani}
\author[aff2,aff4]{Shahriar Shalileh}
\author[aff2,aff3,aff4]{Dena Shahriari}
\cormark[1]
\ead{dena.shahriari@ubc.ca}

\affiliation[aff1]{organization={Department of Electrical and Computer Engineering, The University of British Columbia},
  addressline={5500-2332 Main Mall},
  city={Vancouver},
  state={BC},
  postcode={V6T 1Z4},
  country={Canada}}
\affiliation[aff2]{organization={School of Biomedical Engineering, Biomedical Research Centre, The University of British Columbia},
  addressline={Room 251, 2222 Health Sciences Mall},
  city={Vancouver},
  state={BC},
  postcode={V6T 1Z3},
  country={Canada}}
\affiliation[aff3]{organization={Department of Orthopaedics, Gordon and Leslie Diamond Health Care Centre, The University of British Columbia},
  addressline={11th Floor, 2775 Laurel Street},
  city={Vancouver},
  state={BC},
  postcode={V5Z 1M9},
  country={Canada}}
\affiliation[aff4]{organization={International Collaboration on Repair Discoveries (ICORD), Blusson Spinal Cord Centre},
  addressline={818 West 10th Avenue},
  city={Vancouver},
  state={BC},
  postcode={V5Z 1M9},
  country={Canada}}

\cortext[cor1]{Corresponding author.}

\tnotetext[1]{Author-prepared version of the article published in \textit{Materials \& Design} 268 (2026), 116545. \url{https://doi.org/10.1016/j.matdes.2026.116545}. \textcopyright~2026 The Authors. CC BY 4.0: \url{https://creativecommons.org/licenses/by/4.0/}. Journal backmatter and formatting revised; GitHub link added.}

\begin{abstract}
Kirigami is an increasingly useful fabrication method to produce shape-programmable metamaterial structures. However, inverse design remains difficult because deployment is nonlinear, and feasible cut layouts must satisfy discrete compatibility rules, avoid overlap, and map one target shape to valid designs. We present RL-Kirigami, an inverse design framework that combines optimal-transport conditional flow matching (OT-CFM) with reinforcement learning to generate compatible ratio fields for compact reconfigurable parallelogram quad kirigami. A marching decoder enforces global geometric compatibility, and Group Relative Policy Optimization (GRPO) aligns the generator with nondifferentiable rewards for silhouette matching, feasibility, and ratio-field regularity. Across procedurally generated target shape instances, a single sample from the pretrained OT-CFM prior reached \(94.2\%~\mathrm{sIoU}\) and outperformed solver baselines while reducing forward simulator evaluations from hundreds to \(1\). GRPO improved accuracy to \(94.91\%~\mathrm{sIoU}\) and, with regularity included, reduced \(\mathrm{TV}(\mathbf{x})\) from \(0.95\) to \(0.81\) while maintaining \(94.83\%~\mathrm{sIoU}\). Generated layouts were exported to DXF and laser cut in \(50~\mu\mathrm{m}\) polymeric sheets to produce deployable prototypes in \(8.0\pm 1.0\) minutes per part. These results support a manufacturing-aware inverse design workflow for deployable kirigami metamaterials under geometric feasibility constraints.
Code is available at \url{https://github.com/milad1378yz/RL-Kirigami}.
\end{abstract}


\casnologo
\maketitle
\hypersetup{
  pdfauthor={Milad Yazdani; Shahriar Shalileh; Dena Shahriari},
  pdfsubject={RL-Kirigami inverse silhouette design and laser-cut fabrication}
}

\section{Introduction}
\label{sec:introduction}

Kirigami is the Japanese art of cutting and folding paper sheets. In thin-sheet mechanics, controlled cuts can produce large, controllable deformations~\citep{bertoldi2017flexible}. By designing the layout and geometry of the cut pattern, a flat sheet can open into complex 2D and 3D shapes and achieve prescribed deployed states with controlled stability including bistable or multistable behavior, meaning two or more stable configurations under zero load~\citep{choi2019programming,yang2018multistable}. Furthermore, kirigami cut-pattern design applies across different length scales from atomically thick films to meter scales~\citep{blees2015graphene,wang2022kirigami}.

Combining different features offered by kirigami, the approach has become a widely used fabrication method across different fields~\citep{zhai2021mechanical,wei2024revolutionizing,bliah2025fabrication}. Kirigami also enables functional thin films to accommodate large strain at the microscale~\citep{blees2015graphene}. For example, kirigami patterned skins for soft robots can convert simple inflation or bending into directional friction and motion~\citep{rafsanjani2018kirigami,branyan2022curvilinear,tirado2025multimodal}. In biomedical systems, kirigami patterned implants have been employed for minimally invasive delivery to the body and then expand after placement~\citep{babaee2021kirigami,kim2018shape}. Related slit-pattern design ideas also arise in reconstructive surgery. In meshed split-thickness skin grafting, harvested autologous skin is patterned with rows of slits and expanded to cover larger wounds, with recent work analyzing these expansion limits using a metamaterial-inspired framework~\citep{yu2025meshed}. More broadly, reconstructive procedures such as anterolateral thigh flap reconstruction of large facial skin defects highlight the importance of geometry-aware tissue coverage and shape matching~\citep{mureau2005anterolateral}. For energy applications, kirigami cut sheets can provide integrated solar tracking, and for deployable structures they can fold compactly while deploying to flat surfaces~\citep{lamoureux2015dynamic,wang2022kirigami}. Kirigami cut architectures have also been used to tune wave propagation~\citep{khosravi2022tunable}, electronic response~\citep{won2019stretchable}, and sensing performance~\citep{sun2018kirigami}. Recent reviews of origami and kirigami metamaterials, wearable electronics fabrication, and soft robotics fabrication further show the breadth of applications~\citep{zhai2021mechanical,wei2024revolutionizing,bliah2025fabrication}. Thus, kirigami has become a versatile fabrication methodology. It is also positioned within the broader context of data-driven and programmable metamaterials~\citep{cerniauskas2024machine,zheng2023deep}. Despite the versatility of kirigami, a major consideration remains: configuring a feasible cut pattern for a desired deployed shape or mechanical response.

Solving the kirigami inverse design problem remains a bottleneck~\citep{dudte2023additive,ying2025inverse}. When solving this problem, it is important to account for the deployed response of the full structure that arises from many coupled, almost rigid panels connected by thin ligaments, which are narrow bridges of material left between neighboring cuts. Ligaments can exhibit elastic instabilities as well as strong geometric nonlinearity~\citep{choi2019programming,yang2018multistable,zheng2022continuum}. The response is often related to material properties in addition to the design, sheet thickness, scale, and boundary conditions~\citep{choi2019programming,yang2018multistable,zheng2022continuum}. Even within one tessellation family, small geometric changes can alter stability, load paths, and the deployed configurations that can be reached. As a result, local geometric changes can influence distant regions, which introduces nonlocal coupling. This nonlocal coupling makes manual tuning difficult and rapidly increases the number of design options~\citep{choi2019programming,dudte2023additive}. Continuum and coarse models show that the compatibility limits remain strongly nonlinear even for planar sheets~\citep{zheng2022continuum}. Geometric methods show that the shape of rotating units can control deployed shapes and internal trajectories~\citep{qiao2025inverse}. Practical methods must produce the desired deployed shape under the required boundary conditions and deployability constraints~\citep{dudte2023additive,ying2025inverse}. Furthermore, for shape morphing, the cut pattern must transform from a compact configuration to a deployed one while avoiding overlap and maintaining geometric compatibility~\citep{choi2019programming,dudte2023additive,ying2025inverse}.

Similarly, classical inverse design approaches typically rely on search strategies such as topology optimization, evolutionary algorithms, or particle swarm optimization and use repeated forward simulations to evaluate candidate patterns. These approaches lead to high compute cost per target and require a dedicated optimization loop for each target, which makes generating feasible designs computationally expensive~\citep{xue2017kirigami,perez2007particle,ying2025inverse}. These approaches must also handle nonconvex design spaces and constraints~\citep{perez2007particle,ying2025inverse}.

Data-driven generative models may offer a more effective alternative. More broadly, recent work has shown the promise of data-driven inverse design in programmable and architected metamaterials~\citep{cerniauskas2024machine,zheng2023deep}. After training on simulated or experimental datasets, these models can sample cut patterns conditioned on a prescribed target without iterative search at inference time~\citep{ha2023rapid,chen2025generative}. This can be faster than running an optimizer that calls the forward simulator many times, and these models can represent multiple solution families. Recent work shows inverse mappings for mechanical metamaterials using diffusion and related generative models, including designs that match stress-strain curves and deformation histories~\citep{bastek2023inverse}. GAN-based models have also been used to predict full-field mechanical response in architected metamaterials~\citep{xiang2024gan}. Previous work on kirigami and related architected materials also supports this direction~\citep{alderete2022machine,brzin2025generative}. However, learning kirigami constraints is still challenging. Cut rules are discrete and relational (cuts cannot intersect, and only certain linkages are geometrically possible under compatibility constraints). Deployment must also remain collision-free throughout motion~\citep{choi2019programming,dudte2023additive,ying2025inverse}. Pure feedforward generators can violate hard rules or miss small geometry details that matter for feasibility and matching metrics~\citep{felsch2024generative}. In related mechanical-metamaterial work, guided diffusion can steer generation toward desired mechanical responses~\citep{yang2026guided}. For kirigami, however, such guidance still does not guarantee satisfaction of discrete cut rules; therefore, explicit constraint checks remain useful.

\begin{figure}[pos=t]
    \centering
    \includegraphics[width=\textwidth]{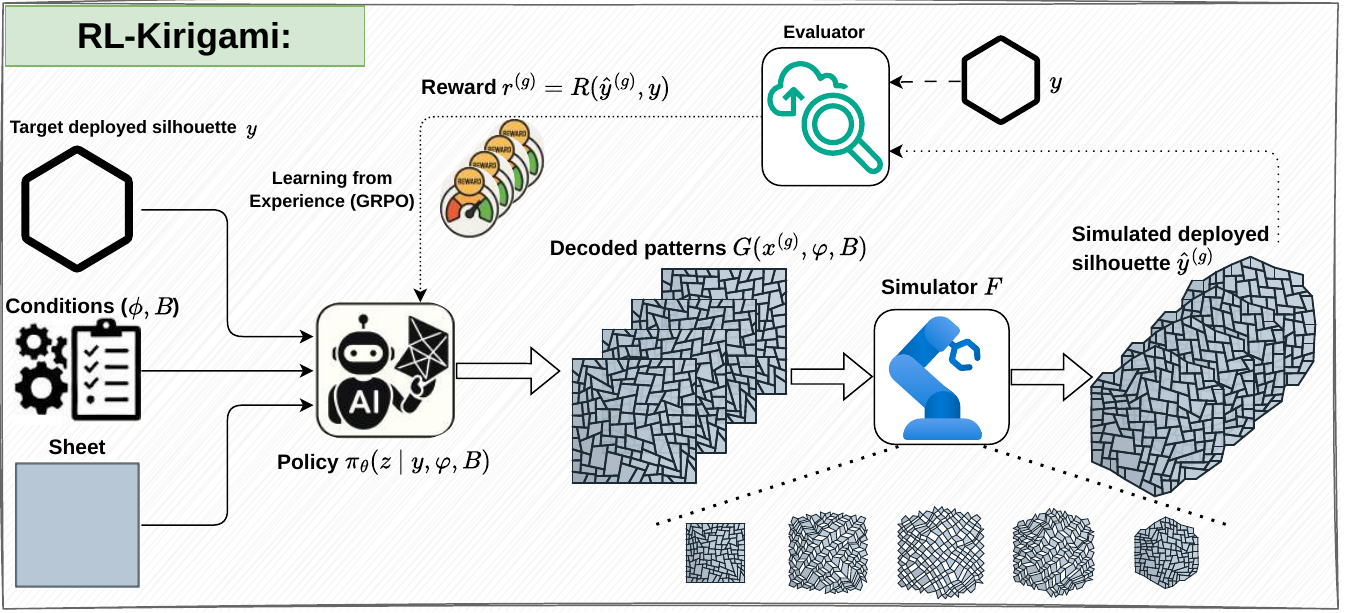}
    \caption{Closed-loop RL-Kirigami fine-tunes the OT-CFM generator with rewards from the forward simulator and evaluator. For a target deployed silhouette \(\mathbf{y}\), the generator proposes a kirigami cut pattern, and the environment simulates \(\widehat{\mathbf{y}}\) for comparison with \(\mathbf{y}\). Group Relative Policy Optimization (GRPO) updates the generator to increase the likelihood of samples that satisfy constraints and match the target.}
    \label{fig:rl-pipeline}
\end{figure}

Building on this direction, RL-Kirigami combines an optimal-transport conditional flow matching (OT-CFM) generator with reinforcement learning fine-tuning under rewards from the forward simulator and evaluator. Flow matching and related continuous time generative models (e.g., rectified flows and stochastic interpolants) learn a velocity field that maps a simple base distribution to the data distribution, and have also been used in other conditional generation settings~\citep{lipman2023flow,liu2023flow,albergo2023building,yazdani2025flow}. This enables fast conditional sampling, which is important because decoding and feasibility checks are required for each candidate. The OT-CFM generator is treated as a stochastic policy over kirigami cut patterns conditioned on the target silhouette. This policy is then refined in a closed loop using reinforcement learning. Rewards combine kirigami validity rules, overlap penalties, and evaluator terms such as shape error and ratio-field regularity computed after forward simulation. This nondifferentiable black-box reward is optimized with Group Relative Policy Optimization (GRPO)~\citep{shao2024deepseekmath}, which aligns the generator with nondifferentiable task rewards and design preferences. Moreover, in structural and metamaterial problems, RL can produce counterintuitive designs and can compete with classical topology optimization when the design space is combinatorial~\citep{brown2022deep,brown2023deep,rosafalco2023reinforcement,shah2021reinforcement}. To our knowledge, the use of RL in kirigami still appears limited and scattered. Our goal here is to open a practical route for inverse kirigami design in which fast conditional sampling handles the one-to-many inverse map, while RL supports preference alignment under discrete feasibility rules and other nondifferentiable or fabrication-oriented preferences without requiring a differentiable surrogate. The marching decoder enforces compatibility within this pipeline. The speed, accuracy, and reliability of each method are compared against solver-based baselines, and the effect of RL on matching performance and ratio-field regularity is evaluated. Finally, we demonstrate the utility of our model by demonstrating an end-to-end laser-cut rapid prototyping workflow.
Fig.~\ref{fig:rl-pipeline} summarizes the RL-Kirigami pipeline.

\section{Methods}
\label{sec:methods}

Fig.~\ref{fig:rl-pipeline} shows the closed-loop inverse design pipeline. Given a target deployed silhouette \(\mathbf{y}\) and conditions (a global deployment parameter \(\varphi\) and boundary anchors \(\mathcal{B}\)), the goal is to find a kirigami design \(\mathbf{x}\) whose decoded deployment matches \(\mathbf{y}\) when passed through the forward simulator. Throughout the paper, \(\mathbf{x}\) denotes the ratio field, \(x_{ij}\) a local ratio, \(\mathbf{y}\) the target silhouette, \(\widehat{\mathbf{y}}\) the simulated deployed silhouette, \(\varphi\) the global deployment parameter, \(\phi_{ij}\) the local cell angle, \(\mathcal{G}\) the marching decoder, and \(\mathcal{F}\) the forward simulator.
Quad kirigami is represented using the negative space (void) view and the local edge/angle notation inspired by Dudte et al.~\citeyearpar{dudte2023additive}. Fig.~\ref{fig:neg-space-geometry}(a) defines the local variables for a void \((i,j)\) and shows how voids tile an \(m\times n\) sheet.
Global consistency is enforced using a marching decoder that rebuilds the full geometry from local ratios and a deployment angle field, together with discrete feasibility checks.
With this representation, RL-Kirigami learns a conditional generative prior for the inverse map by modeling \(P(\mathbf{x}\mid\mathbf{y},\varphi,\mathcal{B})\) using optimal-transport conditional flow matching (OT-CFM)~\citep{lipman2023flow,liu2023flow,albergo2023building}. This prior is then aligned to task rewards using Group Relative Policy Optimization (GRPO)~\citep{shao2024deepseekmath}.

\subsection{Kirigami parameterization and marching decoder}
\label{sec:param_marching}

\begin{figure}[pos=t]
    \centering
    \begin{minipage}[t]{0.62\textwidth}
        \centering
        \makebox[\linewidth][l]{\textbf{(a)}}\\[-0.2em]
        \includegraphics[width=\linewidth]{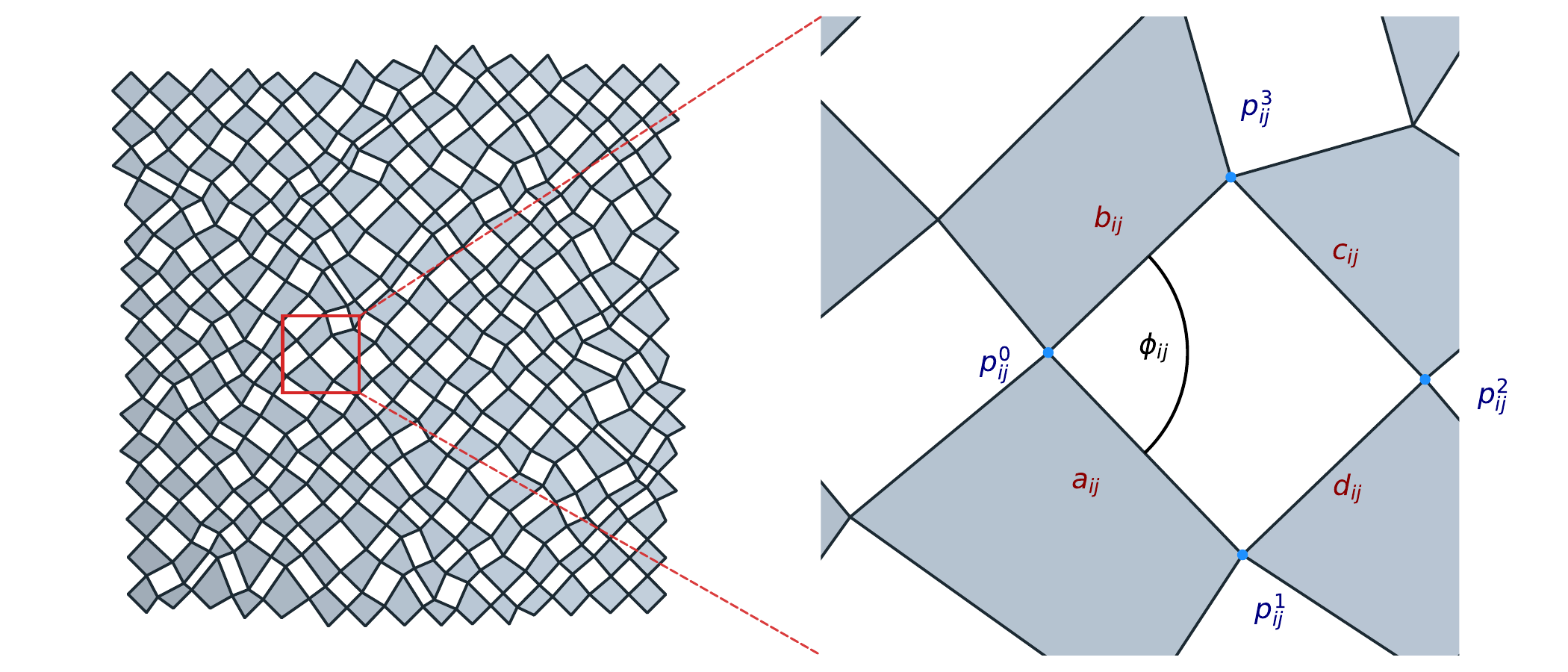}
    \end{minipage}\hfill
    \begin{minipage}[t]{0.34\textwidth}
        \centering
        \makebox[\linewidth][l]{\textbf{(b)}}\\[-0.2em]
        \includegraphics[width=\linewidth]{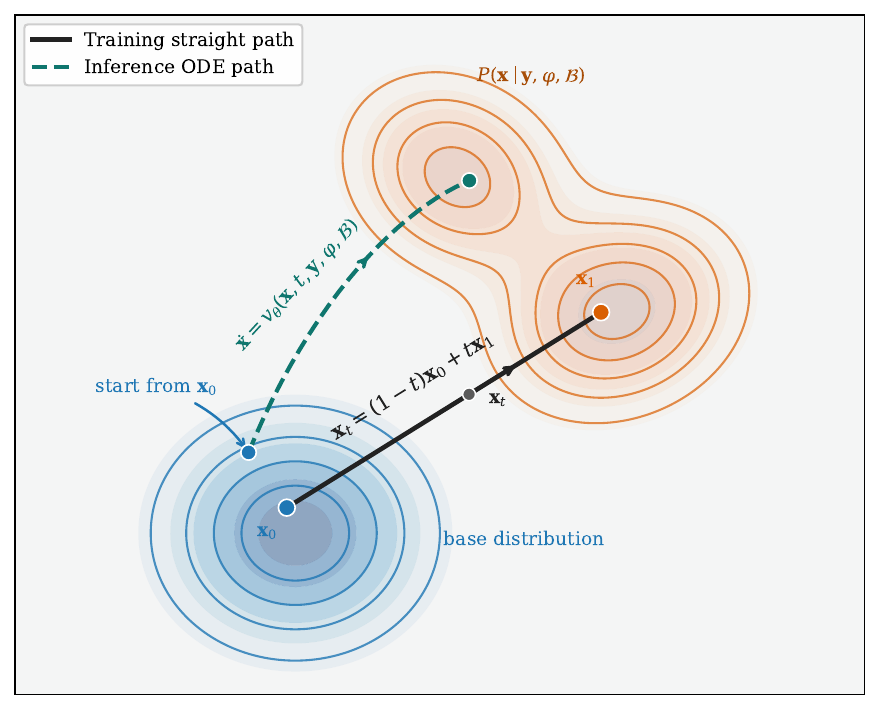}
    \end{minipage}
    \caption{(a) Kirigami sheet and local notation in the negative space view. Left: sheet with an \(m\times n\) array of voids. Right: void \((i,j)\) with corner vertices \(\mathbf{p}_{ij}^{0\ldots 3}\) (blue), side lengths \(a_{ij}\) and \(b_{ij}\), and deployment angle \(\phi_{ij}\) at vertex \(0\), used by the marching decoder (Eq.~\eqref{eq:local_decode}). (b) OT-CFM prior. During training, the model learns the velocity along the straight path from a base sample \(\mathbf{x}_0\) to a feasible design \(\mathbf{x}_1\). At inference, the learned field transports a new base sample toward \(P(\mathbf{x}\mid\mathbf{y},\varphi,\mathcal{B})\).}
    \label{fig:neg-space-geometry}
\end{figure}

A kirigami pattern is modeled as an \(m\times n\) array of voids, as in Fig.~\ref{fig:neg-space-geometry}(a), left. For each cell \((i,j)\), the void is a quadrilateral with corner vertices
\(\mathbf{p}_{ij}^{0\ldots 3}\in\mathbb{R}^2\), indexed counterclockwise, matching Fig.~\ref{fig:neg-space-geometry}(a), right.
The formulation restricts to parallelogram voids and parameterizes each void by an interior deployment angle
\(\phi_{ij}=\angle(\mathbf{p}_{ij}^{1}-\mathbf{p}_{ij}^{0},\,\mathbf{p}_{ij}^{3}-\mathbf{p}_{ij}^{0})\)
and a ratio of side lengths \(x_{ij}=a_{ij}/b_{ij}>0\), where
\(a_{ij}=\|\mathbf{p}_{ij}^{1}-\mathbf{p}_{ij}^{0}\|_2\) and \(b_{ij}=\|\mathbf{p}_{ij}^{3}-\mathbf{p}_{ij}^{0}\|_2\).

Here, deployability is used in a geometric sense: the cut sheet can move from a compact state to the prescribed deployed state while preserving the modeled panel connectivity. For planar quadrilateral kirigami, parallelogram voids provide a constructive route to deployable tessellations, including flat sheets with different topology in the theory of Dang et al.~\citep{dang2021deployable}. The design space is therefore restricted to planar parallelogram voids. The marching decoder enforces this setting by constructing each void as a parallelogram, sharing vertices between neighboring voids, and applying compatibility and overlap checks. Spherical kirigami follows different compatibility conditions~\citep{dang2022spherical} and is outside the present scope.

Only compact reconfigurable parallelogram quad kirigami is considered in this study~\citep{choi2019programming}. In this case, the deployment angle field must follow a checkerboard pattern~\citep{dudte2023additive} and is fully set by a global deployment parameter \(\varphi\):
\begin{equation}
\phi_{ij}=\varphi\quad (i+j\ \text{even}),\qquad
\phi_{ij}=\pi-\varphi\quad (i+j\ \text{odd}).
\label{eq:compact_phi}
\end{equation}

Neighboring voids share corner vertices, so the \(\mathbf{p}_{ij}^{k}\) cannot be chosen separately. The shared corners and consistent tiling are shown in Fig.~\ref{fig:neg-space-geometry}(a), left. Global compatibility is enforced by decoding in a marching order from left to right and top to bottom, reusing vertices that were already created by earlier neighbors. Specifically, when the decoder visits cell \((i,j)\), two vertices are already known from neighboring cells. These known vertices are taken as \(\mathbf{p}_{ij}^{0}\) and \(\mathbf{p}_{ij}^{3}\), consistent with Fig.~\ref{fig:neg-space-geometry}(a), right, and define the seed edge vector
\(\mathbf{s}_{ij}\coloneqq \mathbf{p}_{ij}^{3}-\mathbf{p}_{ij}^{0}\).
Let \(\mathbf{R}(\theta)\) be the planar rotation operator.
Given \(\phi_{ij}\) and \(x_{ij}=a_{ij}/b_{ij}\), the remaining vertices are rebuilt by
\begin{equation}
\mathbf{p}_{ij}^{1}=\mathbf{p}_{ij}^{0}+x_{ij}\,\mathbf{R}(-\phi_{ij})\,\mathbf{s}_{ij},
\qquad
\mathbf{p}_{ij}^{2}=\mathbf{p}_{ij}^{3}+x_{ij}\,\mathbf{R}(-\phi_{ij})\,\mathbf{s}_{ij}.
\label{eq:local_decode}
\end{equation}
This construction yields a parallelogram by design. Once the seed edge length \(b_{ij}=\|\mathbf{s}_{ij}\|_2\) is fixed by already built vertices, \(x_{ij}\) sets the remaining side length via \(a_{ij}=x_{ij}b_{ij}\).

Boundary vertices are initialized along the top and left boundaries of the array, called \(\mathcal{B}\). These anchors fix the global placement and give the starting data for marching. During decoding, the pipeline also computes nondifferentiable geometric checks that act as hard feasibility filters in the forward pipeline: an invalid void count \(N_{\mathrm{inv}}\) for collapsed or self intersecting voids, and an overlap ratio \(r_{\mathrm{ov}}\), defined as \(r_{\mathrm{ov}} = 1 - A_{\mathrm{union}}/\sum_{i=1}^m \sum_{j=1}^n A_{ij}\), where \(A_{ij}\) is the area of void \((i,j)\) and \(A_{\mathrm{union}}\) is estimated from a rasterized union mask of all voids.
If \(N_{\mathrm{inv}}>0\) or \(r_{\mathrm{ov}}>\tau_{\mathrm{ov}}\) for a chosen threshold \(\tau_{\mathrm{ov}}>0\) (or if reconstruction fails), the decode is marked as infeasible.
The pipeline does not differentiate through these checks.

\subsection{Forward simulation and inverse design objective}
\label{sec:designvars}

The independent design variables are the per-void ratios, collected in
\(
\mathbf{x}\equiv[x_{ij}]_{i=1,\ldots,m,\,j=1,\ldots,n}\in\mathbb{R}_+^{m\times n}.
\)
In the experiments, each entry is bounded to a fixed interval to avoid degenerate cases.
The conditioning variables are the fixed boundary anchors \(\mathcal{B}\) and the global deployment parameter \(\varphi\), which sets the local angles through Eq.~\eqref{eq:compact_phi}. The global reconstruction map is
\(
\mathcal{G}:\ (\mathbf{x},\varphi,\mathcal{B})\ \longmapsto\ \{\mathbf{p}_{ij}^{0},\mathbf{p}_{ij}^{1},\mathbf{p}_{ij}^{2},\mathbf{p}_{ij}^{3}\}_{i,j},
\)
implemented by the marching decoder Eq.~\eqref{eq:local_decode} and the discrete checks above. Infeasible decodes are denoted by \(\mathcal{G}(\cdot)=\varnothing\).

Given a decoded pattern, the forward simulator builds the deployed layout at the prescribed state \(\varphi\), recenters the decoded geometry, and rasterizes the union of the decoded quads to a binary silhouette,
\(
\widehat{\mathbf{y}}
=\mathcal{F}\!\left(\mathcal{G}(\mathbf{x},\varphi,\mathcal{B})\right).
\)
In the present experiments, \(\mathcal{F}\) is geometry based: it takes the compatible layout returned by \(\mathcal{G}\), evaluates the deployed shape at the fixed state used for silhouette matching, converts that shape to a mask at fixed resolution, and returns the simulated silhouette. A separate evaluator then compares \(\widehat{\mathbf{y}}\) to the target silhouette \(\mathbf{y}\). In this study, deployed silhouette is used as a task-level structural descriptor of the target shape reached by the kirigami architecture at a prescribed deployment state. In the reported experiments, this comparison uses silhouette IoU with Procrustes alignment~\citep{gower1975generalized} so that the score reflects shape agreement rather than absolute position, rotation, or overall scale. Boundary Chamfer distance is only an optional metric and is not used for the reported reward, model selection, or tables. Additional constraints, such as geometric validity and overlap, are incorporated as penalties in this evaluator. If \(\mathcal{G}(\cdot)=\varnothing\), the sample is treated as infeasible and penalized without calling \(\mathcal{F}\).
The conditional posterior is modeled as
\(
P(\mathbf{x}\mid\mathbf{y},\varphi,\mathcal{B}),
\)
which is usually multimodal because feasibility is discrete, and many different local ratios can lead to the same global shape.

\subsection{OT-CFM prior and GRPO preference alignment}
\label{sec:cfm}

The OT-CFM prior for \(P(\mathbf{x}\mid\mathbf{y},\varphi,\mathcal{B})\) is learned directly on the ratio field \(\mathbf{x}\). Fig.~\ref{fig:neg-space-geometry}(b) gives a simple view of the straight training path and the learned inference path.
Let \(\mathbf{x}_1\) denote a feasible design example, and let \(\mathbf{x}_0\) be a sample from a simple base distribution over ratio fields. The rectified path is
\[
\mathbf{x}_t=(1-t)\mathbf{x}_0+t\mathbf{x}_1,\qquad t\sim\mathcal{U}[0,1].
\]
A velocity field \(v_\theta(\mathbf{x},t,\mathbf{y},\varphi,\mathcal{B})\) is trained with
\begin{equation}
\mathcal{L}_{\mathrm{CFM}}(\theta)=
\mathbb{E}\Big[
\|v_\theta(\mathbf{x}_t,t,\mathbf{y},\varphi,\mathcal{B})-(\mathbf{x}_1-\mathbf{x}_0)\|_2^2
\Big].
\label{eq:cfm_loss}
\end{equation}
This is the standard conditional flow matching objective. During training, optimal transport pairing between base samples and feasible designs is used, which gives the OT-CFM variant used throughout the paper.
At inference time, given \((\mathbf{y},\varphi,\mathcal{B})\), a sample \(\mathbf{x}_0\) is drawn, \(\dot{\mathbf{x}}=v_\theta(\mathbf{x},t,\mathbf{y},\varphi,\mathcal{B})\) is integrated from \(t=0\) to \(t=1\), and the result is decoded using \(\mathcal{G}\).
Conditioning includes \(\mathbf{y}\), \(\varphi\), \(\mathcal{B}\), and positional encodings on the \(m\times n\) grid.

As shown in Fig.~\ref{fig:rl-pipeline}, the generator is fine-tuned using reinforcement learning by treating the flow model as a stochastic policy \(\pi_\theta(\mathbf{x}\mid\mathbf{y},\varphi,\mathcal{B})\), which serves as a trainable approximation to \(P(\mathbf{x}\mid\mathbf{y},\varphi,\mathcal{B})\).
For each target \((\mathbf{y},\varphi,\mathcal{B})\), a group of \(G\) candidates \(\{\mathbf{x}^{(g)}\}_{g=1}^G\sim\pi_\theta(\cdot\mid\mathbf{y},\varphi,\mathcal{B})\) is sampled, decoded with \(\mathcal{G}\), simulated with \(\mathcal{F}\), and assigned rewards
\[
\widehat{\mathbf{y}}^{(g)}=\mathcal{F}(\mathcal{G}(\mathbf{x}^{(g)},\varphi,\mathcal{B})),
\qquad
r^{(g)}=R\!\left(\mathbf{x}^{(g)},\widehat{\mathbf{y}}^{(g)},\mathbf{y}\right).
\]
Here \(R\) is a scalar reward computed by the evaluator, where higher is better. It may depend on the sampled ratio field, the simulated silhouette, and decoded-geometry penalties such as overlap or ratio-field regularity. If \(\mathcal{G}(\cdot)=\varnothing\) due to the discrete checks in Sec.~\ref{sec:param_marching}, the sample is treated as infeasible and assigned a fixed low reward without calling \(\mathcal{F}\).
The parameters \(\theta\) are updated using GRPO-style group relative reward weighting. Let \(\mu\) and \(\sigma\) be the mean and standard deviation of \(\{r^{(g)}\}\) inside a group, and let \(\varepsilon>0\) be a small constant:
\begin{equation}
A^{(g)}=\frac{r^{(g)}-\mu}{\sigma+\varepsilon}.
\label{eq:grpo_adv}
\end{equation}
These relative advantages are turned into positive weights with temperature \(T>0\):
\begin{equation}
w^{(g)}=\frac{\exp(A^{(g)}/T)}{\frac{1}{G}\sum_{k=1}^G\exp(A^{(k)}/T)},
\label{eq:grpo_weights}
\end{equation}
and GRPO is applied by reweighting the flow matching regression objective toward higher reward samples:
\begin{equation}
\mathcal{L}_{\mathrm{GRPO}}(\theta)=
\mathbb{E}\Bigg[
\frac{1}{G}\sum_{g=1}^G
w^{(g)}\,\ell_{\mathrm{CFM}}\!\left(\theta;\mathbf{x}^{(g)},\mathbf{y},\varphi,\mathcal{B}\right)
\Bigg],
\label{eq:grpo_loss}
\end{equation}
where \(\ell_{\mathrm{CFM}}\) is the per sample squared error term inside Eq.~\eqref{eq:cfm_loss}. In simple terms, this update supports preference alignment by making the model sample high reward designs more often, while keeping the basic structure learned by the OT-CFM prior.
During training and evaluation, \(\mathcal{G}\) is always the marching decoder in Eq.~\eqref{eq:local_decode}. A global design matrix reconstruction is not used.

\section{Experimental Results}
\label{sec:experiments}

The experiments address four questions: \textbf{(i)} Can the OT-CFM prior used in RL-Kirigami solve the kirigami inverse design problem reliably under the given conditions, and how does it compare to solver-based inverse design in speed, accuracy, and robustness? \textbf{(ii)} Among common conditional generative models, which one is the best choice for this inverse design task under the same representation and the same geometric decoder? \textbf{(iii)} Can reinforcement learning in RL-Kirigami improve task accuracy and ratio-field regularity through preference alignment? \textbf{(iv)} Can the resulting designs be fabricated rapidly in situ using laser cutting on polyamide, and do representative prototypes show the intended deployed behavior? Unless stated otherwise, all methods use the same kirigami parameterization and marching decoder (Sec.~\ref{sec:param_marching}), the same global deployment parameterization (Eq.~\eqref{eq:compact_phi}), and the same feasibility checks including invalid void checks and the overlap threshold \(\tau_{\mathrm{ov}}\).

\subsection{Overall experimental setup}
\label{subsec:exp_setup}

A common setup is used across the computational inverse-design experiments. The task is single-state silhouette matching at a fixed deployment parameter \(\varphi\) using the ratio field \(\mathbf{x}\) from Sec.~\ref{sec:designvars}. A fixed \(10\times 10\) grid is used, and target silhouettes \(\mathbf{y}\) are rendered as binary masks at \(128\times 128\) resolution.
Candidate pairs \((\mathbf{y},\mathbf{x})\) are generated using Sobol quasi-random sampling on \([-1,1]^{10\times 10}\) for more even coverage of the ratio-field space, then mapping each entry through \(x_{ij}=10^{z_{ij}}\), which gives the ratio-field range \([1/10,10]\). Infeasible candidates are discarded before rasterizing the deployed silhouette to obtain \(\mathbf{y}\) (details in Appendix~\ref{supp:data_generation}). One train/validation/test split is used for the reported training and evaluation runs, with \(5000\) feasible training samples, \(500\) feasible validation samples, and \(500\) feasible test samples. Unless stated otherwise, reported evaluation runs use the full test split. All training, inference, and solver runs are executed on the same workstation. Hardware details are listed in Appendix~\ref{supp:hardware}, and detailed data generation and hyperparameter settings are listed in Appendices~\ref{supp:data_generation} and~\ref{supp:hyperparams}.
The solver, generator, and RL benchmarks use feasible targets from this ratio-field generator. The OOD robustness test in Sec.~\ref{exp:ood_silhouette} uses target masks generated independently of that process to probe distribution shift.
Unless stated otherwise, one candidate design is drawn per target. Reported quantitative results are averaged over evaluation targets. Values written as \(a \pm b\) report the mean and standard deviation over three evaluation seeds \(\{0,1,2\}\). Other reported metrics are means over the same evaluation seeds.

\subsection{Flow matching vs.\ solver-based inverse design}
\label{exp:fm_vs_solver}

\begin{figure}[pos=tbp]
    \centering
    \includegraphics[width=0.96\textwidth]{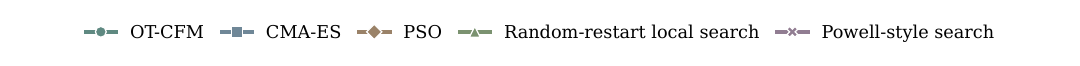}\\[-0.2em]
    \begin{minipage}[t]{0.48\textwidth}
        \centering
        \makebox[\linewidth][l]{\textbf{(a)}}\\[0.25em]
        \includegraphics[width=\linewidth]{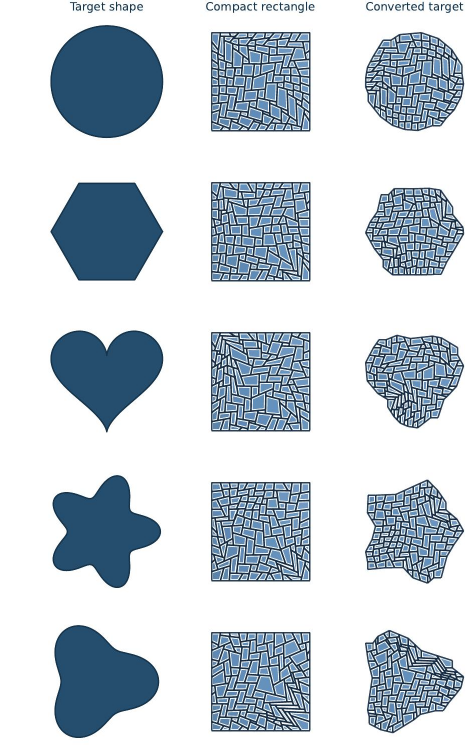}
    \end{minipage}\hfill
    \begin{minipage}[t]{0.49\textwidth}
        \centering
        \makebox[\linewidth][l]{\textbf{(b)}}\\[-0.3em]
        \includegraphics[width=0.94\linewidth]{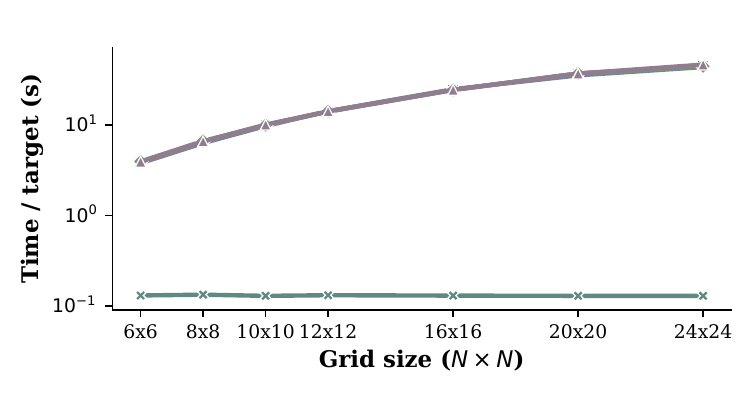}\\[-0.1em]
        \makebox[\linewidth][l]{\textbf{(c)}}\\[-0.3em]
        \includegraphics[width=0.94\linewidth]{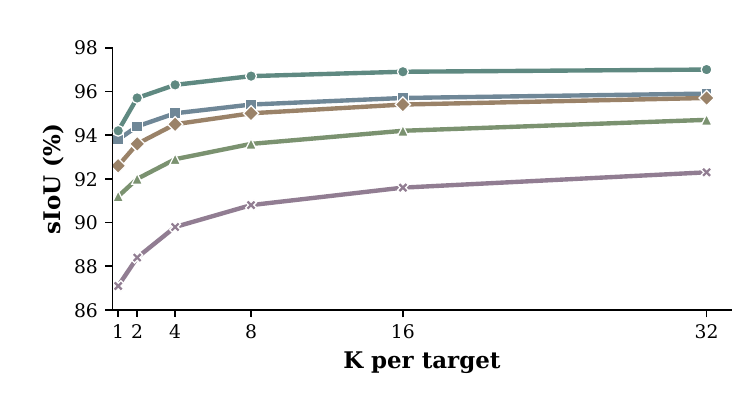}\\[-0.1em]
        \makebox[\linewidth][l]{\textbf{(d)}}\\[-0.3em]
        \includegraphics[width=0.94\linewidth]{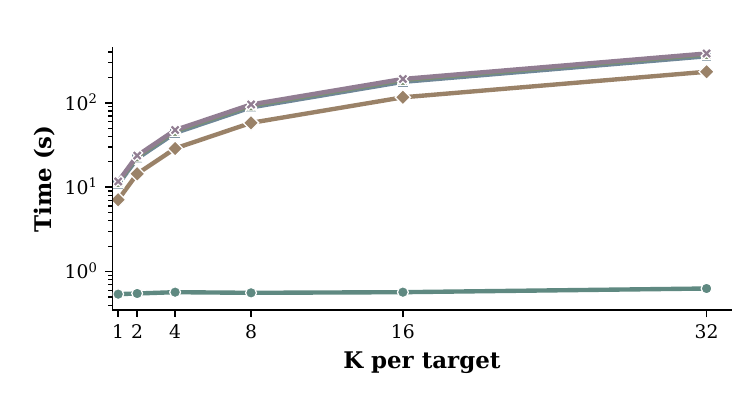}
    \end{minipage}
    \caption{Inverse-design target examples and solver comparisons. (a) Representative target silhouettes, compact rectangle, and converted target layouts. (b) Wall clock time vs.\ grid size for the solver experiment. (c,d) Best-of-\(K\) sensitivity on the full test split. Panel (c) reports mean \(\mathrm{sIoU}\) vs.\ \(K\). Panel (d) reports total wall clock time per target on a log scale.}
    \label{fig:inverse_design_summary}
\end{figure}

Evaluation uses the shared test split. The pretrained OT-CFM prior is sampled once per target (\(K=1\)) using Euler integration with \(8\) steps. Table~\ref{tab:fm_vs_solver} reports Procrustes-aligned silhouette intersection over union (\(\mathrm{sIoU}\))~\citep{gower1975generalized}, success rate, and forward simulator evaluations as \(\#\mathcal{F}\). Success is defined as \(\mathrm{sIoU}\ge \tau_{\mathrm{sIoU}}\) with no feasibility violations under the checks in Sec.~\ref{sec:param_marching}, with \(\tau_{\mathrm{sIoU}}=0.85\) and \(\tau_{\mathrm{ov}}=0.02\). The exact \(\mathrm{sIoU}\) calculation is summarized in Appendix~\ref{supp:siou_metric}. Solver methods use tolerance-based stopping with solver x-tolerance \(10^{-3}\), relative objective tolerance \(10^{-3}\), patience \(5\), and a \(1000\)-evaluation safety cap per target. No feasibility violations were observed in this setting. OT-CFM is compared with representative solver-based inverse design methods that optimize the ratio field \(\mathbf{x}\) under the same decoder \(\mathcal{G}\): Covariance Matrix Adaptation Evolution Strategy (CMA-ES)~\citep{hansen2001completely}, particle swarm optimization (PSO)~\citep{perez2007particle}, random-restart local search~\citep{boender1982stochastic}, and a bounded Powell-style nonlinear search inspired by Dudte et al.~\citeyearpar{dudte2023additive}. Solver-specific search settings are summarized in Appendix~\ref{supp:hyperparams} and Table~\ref{tab:supp_hyperparams}. In Table~\ref{tab:fm_vs_solver}, OT-CFM reaches \(94.2\%\) \(\mathrm{sIoU}\) and \(98.7\%\) success with one forward evaluation per target, while the best solver baseline reaches \(93.8\%\) \(\mathrm{sIoU}\) and the solver baselines use \(625\) to \(1000\) evaluations per target.

Fig.~\ref{fig:inverse_design_summary}(a) shows representative target silhouettes, including extreme cases, together with the compact rectangle and converted target layouts.

\begin{table}[pos=t]
    \centering
    \caption{Comparison of the OT-CFM prior (\(K=1\)) with solver-based baselines for silhouette matching on the full test split. Columns report \(\mathrm{sIoU}\), success rate (\(p_{\mathrm{succ}}\)), and forward simulator evaluations \(\#\mathcal{F}\). Solver baselines use the same tolerance-based stopping described in Sec.~\ref{exp:fm_vs_solver}.}
    \label{tab:fm_vs_solver}
    \setlength{\tabcolsep}{3pt}
    \begin{tabular}{@{}p{6.8cm}ccc@{}}
        \hline
        Method & \(\mathrm{sIoU}\uparrow\) (\%) & \(p_{\mathrm{succ}}\uparrow\) (\%) & \(\#\mathcal{F}\downarrow\) \\
        \hline
        CMA-ES~\citep{hansen2001completely} & \(\meanstd{93.8}{2.9}\) & \(97.0\)  & \(986.0\) \\
        PSO~\citep{perez2007particle} & \(\meanstd{92.6}{3.3}\) & \(97.2\) & \(625.1\) \\
        Random-restart local search~\citep{boender1982stochastic} & \(\meanstd{91.2}{4.9}\) & \(89.8\) & \(1000.0\) \\
        \makecell[l]{Bounded Powell-style nonlinear search\\ inspired by Dudte et al.~\citeyearpar{dudte2023additive}} & \(\meanstd{87.1}{5.9}\) & \(76.0\) & \(1000.0\) \\ 
        \hline
        OT-CFM & \(\meanstd{94.2}{0.7}\) & \(98.7\) & \(1.0\) \\
        \hline
    \end{tabular}
\end{table}

Fig.~\ref{fig:inverse_design_summary}(b) compares per-target wall-clock time as the square grid size changes from \(6\times 6\) to \(24\times 24\). In this sweep, the ratio field has \(N^2\) free variables, increasing the search dimension from \(36\) to \(576\) parameters. Solver baselines use the same tolerance-based setup as Table~\ref{tab:fm_vs_solver}, averaged over three representative \(128\times 128\) target masks (Heart, Circle, and Hexagon). The OT-CFM curve reports one Euler-\(8\) sample plus evaluation per target at each grid size, with the corresponding prior retrained at that resolution. On the log-scale timing plot, solver time grows from a few seconds at \(6\times 6\) to a few tens of seconds at \(24\times 24\), while OT-CFM stays around \(0.14\)~s.

For the finite sheet setting used here, grid size is treated as a design resolution parameter rather than a universal representative length scale. Appendix~\ref{supp:design_resolution_ablation} and Table~\ref{tab:supp_design_resolution_ablation} give the \mbox{\(6\times 6\)} to \mbox{\(12\times 12\)} ablation used to select \mbox{\(10\times 10\)}, while Fig.~\ref{fig:inverse_design_summary}(b) reports timing up to \mbox{\(24\times 24\)}.

Fig.~\ref{fig:inverse_design_summary}(c,d) gives best-of-\(K\) sensitivity on the same full test split. For the OT-CFM prior, \(K\) is the number of generated candidates per target; for solver baselines, \(K\) is the number of independent runs whose best final result is retained. Both families are shown up to \(K=32\). Larger \(K\) improves each retained result, but the generator remains much faster: at the shared \(K=4\) setting, OT-CFM is slightly above \(96\%\) \(\mathrm{sIoU}\) and below \(1\)~s per target, whereas PSO is slightly below \(95\%\) and already in the tens of seconds.

\subsection[Out-of-distribution silhouette robustness evaluation]{Out-of-distribution (OOD) silhouette robustness evaluation}
\label{exp:ood_silhouette}

To complement the benchmark in Sec.~\ref{exp:fm_vs_solver}, an OOD target set was constructed from a shape process independent of the kirigami ratio-field generator. The set contains \mbox{\(37\)} target silhouettes at the same \mbox{\(128\times 128\)} resolution and uses the same decoder, simulator, and \mbox{\(\mathrm{sIoU}\)} evaluator as the main experiments. Each target is assigned to one of four groups using measured mask geometry. A small set of literal targets, such as letters and a doodle, is kept as the literal group. The other targets are grouped by geometry: targets with an interior hole are in the group with holes, targets with solidity at least \mbox{\(0.98\)} are convex, and the remaining targets are concave. Solidity is the mask area divided by its convex hull area. To quantify distribution shift, principal component analysis (PCA) is fit to vectorized training masks~\citep{jolliffe2016principal}. Two OOD scores are then computed: distance to the nearest training mask in PCA score space and PCA reconstruction error after projection to the same components. The validation split sets the reference scale for the training distribution.

For each OOD target, one OT-CFM sample is compared with a Powell search baseline inspired by Dudte et al.~\citeyearpar{dudte2023additive}. The evaluation reports Procrustes aligned \mbox{\(\mathrm{sIoU}\)}, feasibility status, overlap ratio, and failure category. The cut topology is kept fixed for targets with holes. These cases therefore test the boundary of the current compact parallelogram quad kirigami family, rather than hole design with a changed topology.

\begin{figure}[pos=!tbp]
    \centering
    \includegraphics[width=\textwidth]{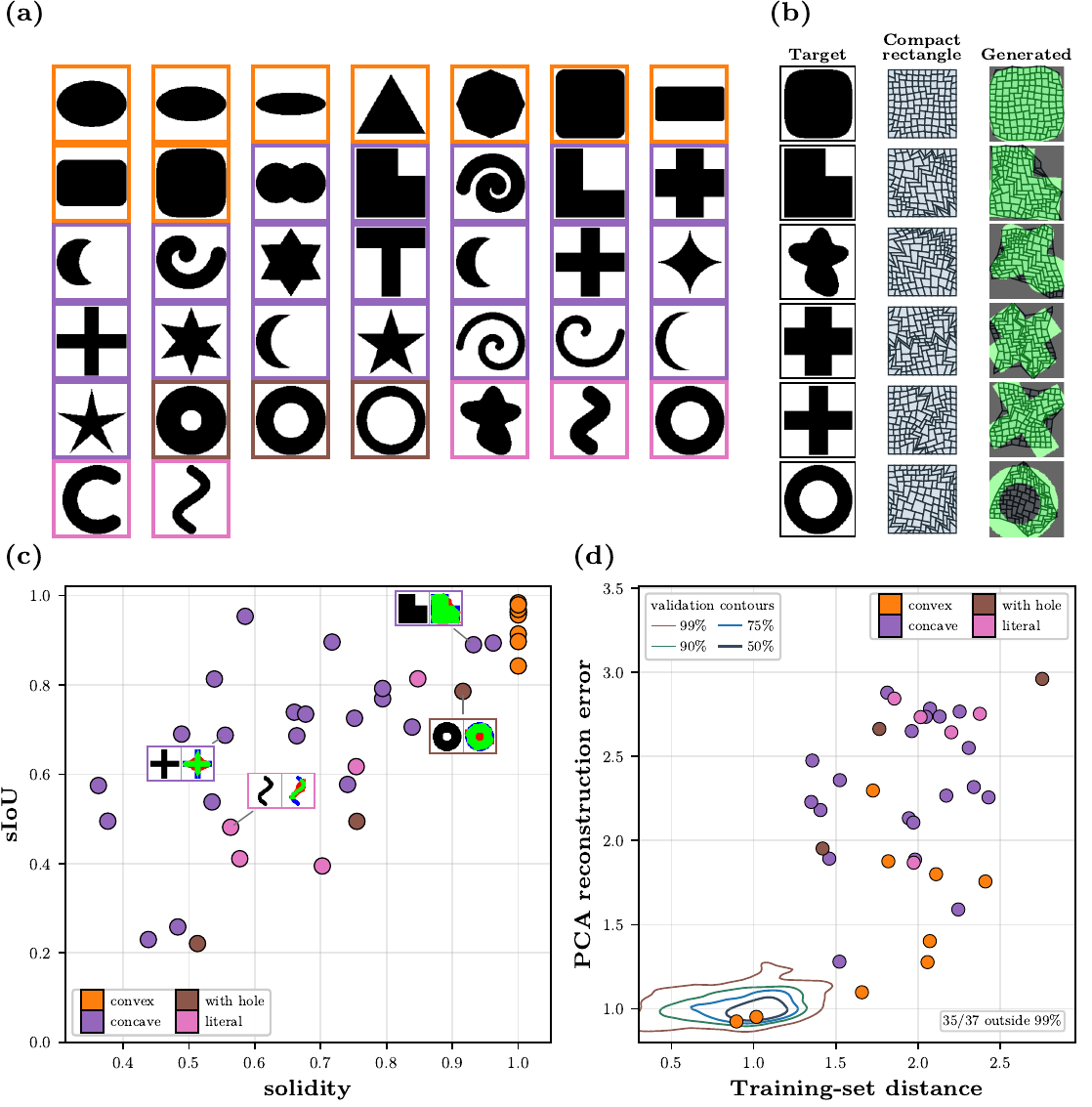}
    \caption{Out-of-distribution silhouette robustness evaluation. (a) The \mbox{\(37\)} OOD targets, with border color indicating group. (b) Representative cases shown as target, compact rectangle, and generated structure with the aligned target overlaid. (c) Procrustes aligned \mbox{\(\mathrm{sIoU}\)} against solidity for every target, colored by group, with selected target and overlay callouts. (d) PCA OOD score view. The horizontal axis is nearest training mask distance in the PCA score space with \mbox{\(64\)} components, normalized by the validation median. The vertical axis is PCA reconstruction error with the same \mbox{\(64\)} components, normalized by the validation median. Contours show validation density, and the contour legend gives the approximate fraction of validation masks enclosed by each contour.}
    \label{fig:ood_overview}
\end{figure}

Fig.~\ref{fig:ood_overview} shows the OOD target set colored by group in panel (a), representative cases in panel (b), \mbox{\(\mathrm{sIoU}\)} against solidity in panel (c), and the PCA OOD score view in panel (d). The first \mbox{\(64\)} PCA components explain \mbox{\(82.3\%\)} of the training mask variance. In this PCA space, the OOD targets have median nearest training mask distance \mbox{\(1.97\times\)} the validation median and median reconstruction error \mbox{\(2.26\times\)} the validation median. As an additional nearest neighbor mask check, their median nearest training mask IoU is \mbox{\(56.2\%\)}, compared with \mbox{\(90.8\%\)} for validation masks. In the PCA OOD score space, \mbox{\(35\)} of the \mbox{\(37\)} OOD targets are outside the \mbox{\(99\%\)} validation contour.

Across the \mbox{\(37\)} OOD targets, the OT-CFM prior reaches mean \mbox{\(\mathrm{sIoU}\)} \mbox{\(71.2\%\)}, while the solver baseline reaches \mbox{\(62.0\%\)}. The OT-CFM prior gives higher \mbox{\(\mathrm{sIoU}\)} on \mbox{\(28\)} targets. Table~\ref{tab:ood_group_summary} summarizes the group level scores. For \mbox{\(3\)} targets, the decoded quads overlap above the \mbox{\(0.02\)} feasibility cutoff and have invalid quadrilaterals, so the feasibility filter from Sec.~\ref{sec:param_marching} rejects them. For the remaining feasible cases, lower scores indicate silhouette mismatch rather than measured feasibility failure.

\begin{table}[pos=t]
    \centering
    \caption{OOD silhouette scores by group. The OT-CFM prior uses one sample per target. The solver uses the same decoder and feasibility checks.}
    \label{tab:ood_group_summary}
    \footnotesize
    \setlength{\tabcolsep}{3pt}
    \renewcommand{\arraystretch}{1.08}
    \begin{tabular}{@{}lcc@{}}
        \hline
        Group & OT-CFM \mbox{\(\mathrm{sIoU}\)} (\%) & Solver \mbox{\(\mathrm{sIoU}\)} (\%) \\
        \hline
        Convex & 94.0 & 81.2 \\
        Concave & 68.2 & 59.4 \\
        Literal & 54.3 & 46.7 \\
        With holes & 50.0 & 47.2 \\
        \hline
    \end{tabular}
\end{table}

Table~\ref{tab:ood_group_summary} shows that convex OOD targets remain close to the score in Table~\ref{tab:fm_vs_solver}, \mbox{\(94.0\%\)} compared with \mbox{\(94.2\%\)}. This suggests that they are mostly compatible with the current family. Relative to the convex group, OT-CFM drops by \mbox{\(25.8\)}, \mbox{\(39.7\)}, and \mbox{\(44.0\)} points for concave, literal, and targets with holes. The concave group is limited mainly by boundary concavity. Literal targets are harder because symbolic shapes must be matched with the same compact family. Targets with holes remain poorly matched under the fixed topology. These drops show the main boundary observed within the OOD set. Representative failure cases are shown in Fig.~\ref{fig:supp_failure_cases}.

\subsection{Conditional generative model comparison for inverse design}
\label{exp:gen_model_choice}

Table~\ref{tab:gen_models} compares cVAE, cGAN, diffusion, and OT-CFM. OT-CFM reaches \(94.2\%\) \(\mathrm{sIoU}\) and \(98.7\%\) success. Diffusion reaches the highest IS at \(2.05\), with a sampling time of \(473.30\)~ms compared with \(159.64\)~ms for OT-CFM.

\begin{table}[pos=t]
    \centering
    \caption{Comparison of conditional generators under the same representation and geometric decoder. Accuracy, success, and generator sampling time use one sample per target (\(K=1\)). IS uses \(K=16\) conditional samples per target~\citep{salimans2016improved}.}
    \label{tab:gen_models}
    \begin{tabular}{lcccc}
        \hline
        Model & \(\mathrm{sIoU}\uparrow\) (\%) & \(p_{\mathrm{succ}}\uparrow\) (\%) & Time/sample (ms)\(\downarrow\) & IS\(\uparrow\) \\
        \hline
        cVAE & \(\meanstd{83.6}{6.0}\) & \(51.0\) & \(14.06\) & \(1.62\) \\
        cGAN & \(\meanstd{87.6}{0.3}\) & \(77.8\) & \(20.62\) & \(1.74\) \\
        Diffusion & \(\meanstd{85.5}{0.3}\) & \(76.2\) & \(473.30\) & \(2.05\)  \\
        OT-CFM & \(\meanstd{94.2}{0.7}\) & \(98.7\) & \(159.64\) & \(1.89\)\\
        \hline
    \end{tabular}
\end{table}

\begin{figure}[pos=t]
    \centering
    \includegraphics[width=0.86\textwidth]{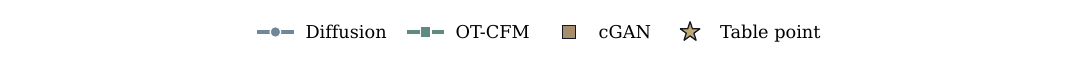}\\[-0.2em]
    \begin{minipage}[t]{0.48\textwidth}
        \centering
        \makebox[\linewidth][l]{\textbf{(a)}}\\[-0.25em]
        \includegraphics[width=0.9\linewidth]{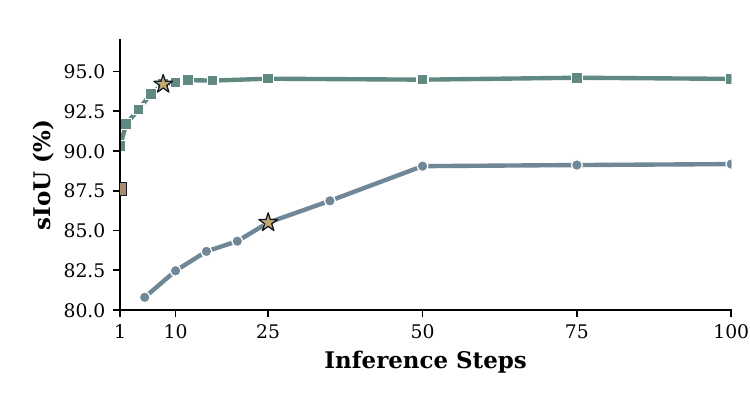}
    \end{minipage}\hfill
    \begin{minipage}[t]{0.45\textwidth}
        \centering
        \makebox[\linewidth][l]{\textbf{(b)}}\\[-0.25em]
        \includegraphics[width=\linewidth]{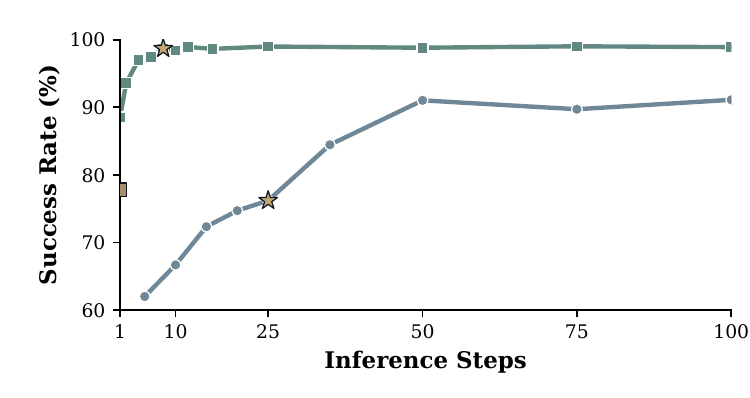}
    \end{minipage}\\[0.8em]
    \begin{minipage}{0.8\textwidth}
        \centering
        \makebox[\linewidth][l]{\textbf{(c)}}\\[-0.25em]
        \includegraphics[width=\linewidth]{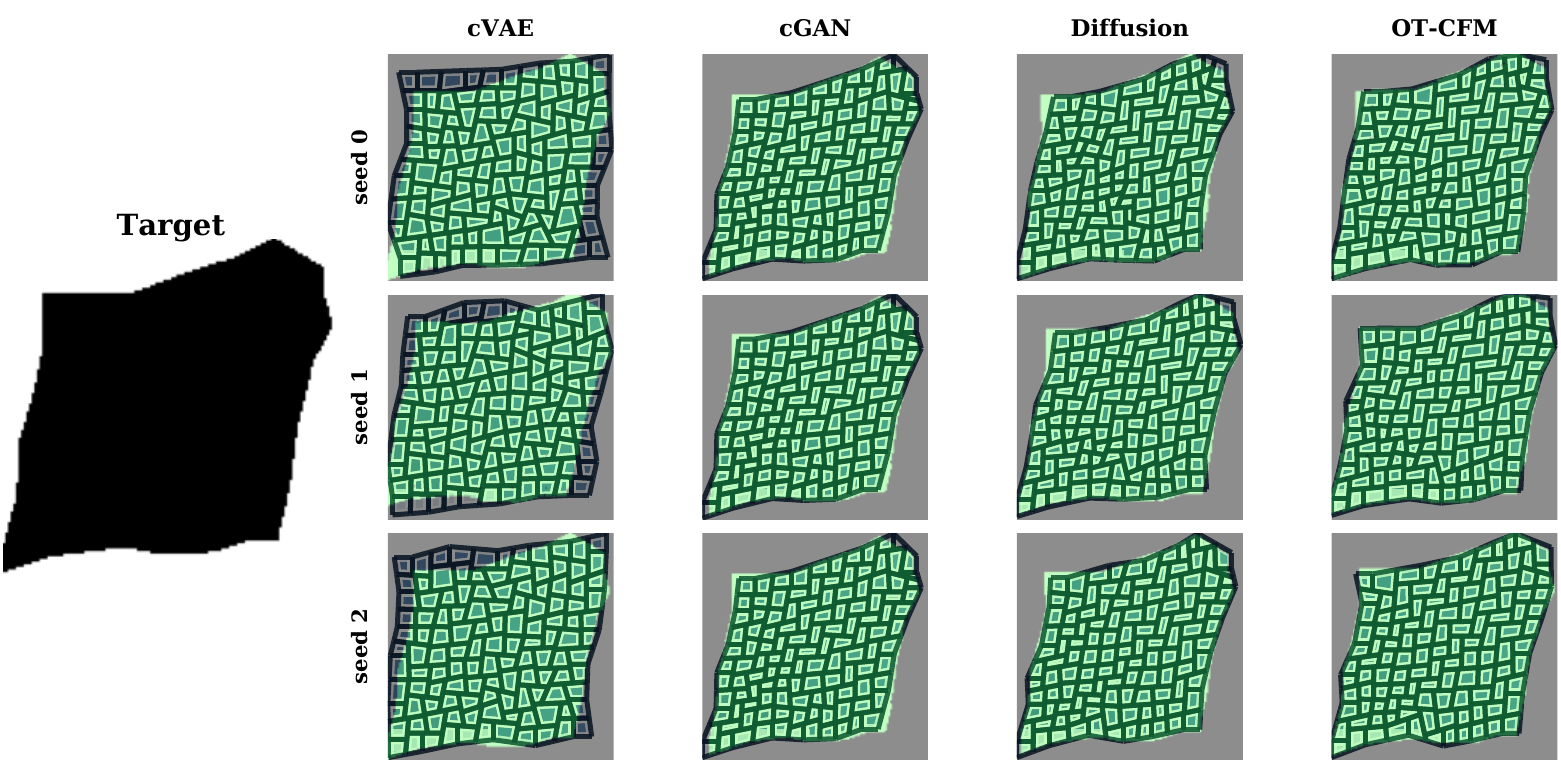}
    \end{minipage}
    \caption{Conditional generator comparison. (a) \(\mathrm{sIoU}\) vs.\ inference steps. (b) Success rate vs.\ inference steps. Star markers show the reported settings. Square markers show the \(1\)-step cGAN baseline. (c) Three-sample visual comparison for the conditional generators in Table~\ref{tab:gen_models} on one randomly selected target from the validation split. Rows show three independent samples.}
    \label{fig:gen_steps_sensitivity}
\end{figure}

Fig.~\ref{fig:gen_steps_sensitivity}(a,b) shows the sensitivity to the number of sampling steps. At \(2\) steps, OT-CFM reaches \(91.7\%\) \(\mathrm{sIoU}\) and \(94.0\%\) success, while diffusion reaches \(89.2\%\) and \(90.7\%\) at \(100\) steps.

Fig.~\ref{fig:gen_steps_sensitivity}(c) shows one randomly selected target from the validation split with three independent \(K=1\) samples for one target. On this example, OT-CFM reaches mean \(\mathrm{sIoU}=96.4\%\) across the three samples, compared with \(92.4\%\) for diffusion, \(90.4\%\) for cGAN, and \mbox{\(89.2\%\)} for cVAE.

\subsection{RL fine-tuning for accuracy and regularity}
\label{exp:rl_acc_phys}

RL-Kirigami is initialized from the OT-CFM prior selected in Sec.~\ref{exp:gen_model_choice}. GRPO fine-tuning uses the training split, with validation used only for model selection, and the final reported results are evaluated on the shared test split with Euler-\(8\) sampling. This section studies GRPO preference alignment for accuracy and ratio-field regularity, measured by \(\mathrm{sIoU}\) and \(\mathrm{TV}(\mathbf{x})\). The OT-CFM prior without RL is compared with three GRPO variants: accuracy only, regularity only, and accuracy plus regularity.

\subsubsection{(A) RL for accuracy}
The accuracy-only RL reward is
\[
R(\mathbf{x},\widehat{\mathbf{y}},\mathbf{y})=\mathrm{sIoU}
-5.0\,\mathbf{1}[\mathcal{G}(\cdot)=\varnothing]
-2.0\,\mathbf{1}[N_{\mathrm{inv}}>0]
-2.0\,[r_{\mathrm{ov}}-\tau_{\mathrm{ov}}]_+,
\]
where \(N_{\mathrm{inv}}\) is the invalid-quadrilateral count from Sec.~\ref{sec:param_marching}, \(r_{\mathrm{ov}}\) is the overlap ratio, and \([u]_+\coloneqq\max(u,0)\). Thus, reconstruction failure uses penalty \(5.0\), invalid-quadrilateral presence uses penalty \(2.0\), and overlap beyond \(\tau_{\mathrm{ov}}\) uses weight \(2.0\).
A fixed GRPO group size \(G=4\) is used, and results are reported against total environment calls, where one call means one sampled candidate passed through decoding, feasibility checks, and, if feasible, forward simulation followed by evaluation. The OT-CFM sampling setup from Sec.~\ref{exp:gen_model_choice} is kept during RL-Kirigami rollouts and during the final evaluation.

\subsubsection{(B) RL for physical regularity (ratio-field regularity)}
To encode a regularity preference, a total variation (TV) term is added to penalize the spatial variation of \(\mathbf{x}\):
\[
\mathrm{TV}(\mathbf{x})=
\sum_{i=1}^{m-1}\sum_{j=1}^{n}\big|x_{i+1,j}-x_{i,j}\big|
\;+\sum_{i=1}^{m}\sum_{j=1}^{n-1}\big|x_{i,j+1}-x_{i,j}\big|,
\]
and the augmented reward \(R(\mathbf{x},\widehat{\mathbf{y}},\mathbf{y})-\lambda_{\mathrm{tv}}\;\mathrm{TV}(\mathbf{x})\) is optimized, where \(\lambda_{\mathrm{tv}}\ge 0\) sets the weight on the TV term. For the regularity-only variant, the reward omits the \(\mathrm{sIoU}\) term and keeps the same feasibility penalties together with the TV term. For the accuracy plus regularity variant, the normalized accuracy and regularity rewards contribute equally, so half of the hybrid reward comes from each term. The results below report three RL settings at a common budget of \(10000\) environment calls: accuracy only, regularity only, and accuracy plus regularity.

Table~\ref{tab:rl_results_new} reports the \(10000\)-call values for each RL variant. Accuracy-only GRPO gives the highest \(\mathrm{sIoU}\) at \(94.91\%\), while its \(\mathrm{TV}(\mathbf{x})\) is \(0.981\). Regularity-only GRPO gives the lowest \(\mathrm{TV}(\mathbf{x})\) at \(0.60\), with lower \(\mathrm{sIoU}\) at \(93.67\%\). The combined reward keeps \(\mathrm{sIoU}\) high at \(94.83\%\) and lowers \(\mathrm{TV}(\mathbf{x})\) to \(0.81\).

\begin{table}[pos=t]
    \centering
    \caption{Effect of GRPO preference alignment on accuracy and ratio-field regularity. Columns report \(\mathrm{sIoU}\) and total variation \(\mathrm{TV}(\mathbf{x})\) (lower means higher regularity).}
    \label{tab:rl_results_new}
    \setlength{\tabcolsep}{3pt}
    \begin{tabular}{@{}lcc@{}}
        \hline
        Method & \(\mathrm{sIoU}\uparrow\) (\%) & \(\mathrm{TV}(\mathbf{x})\downarrow\) \\
        \hline
        OT-CFM & \(\meanstd{94.2}{0.7}\) & \(\meanstd{0.95}{0.10}\) \\
        RL-Kirigami (accuracy) & \(\meanstd{94.91}{0.1}\) & \(\meanstd{0.98}{0.11}\) \\
        RL-Kirigami (regularity) & \(\meanstd{93.67}{0.2}\) & \(\meanstd{0.60}{0.12}\) \\
        RL-Kirigami (accuracy + regularity) & \(\meanstd{94.83}{0.1}\) & \(\meanstd{0.81}{0.17}\) \\
        \hline
    \end{tabular}
\end{table}

\begin{figure}[pos=!thb]
    \centering
    \includegraphics[width=0.9\textwidth]{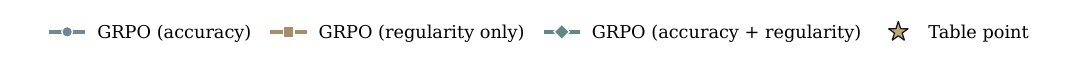}\\[-0.2em]
    \begin{minipage}[t]{0.48\textwidth}
        \centering
        \makebox[\linewidth][l]{\textbf{(a)}}\\[-0.25em]
        \includegraphics[width=\linewidth]{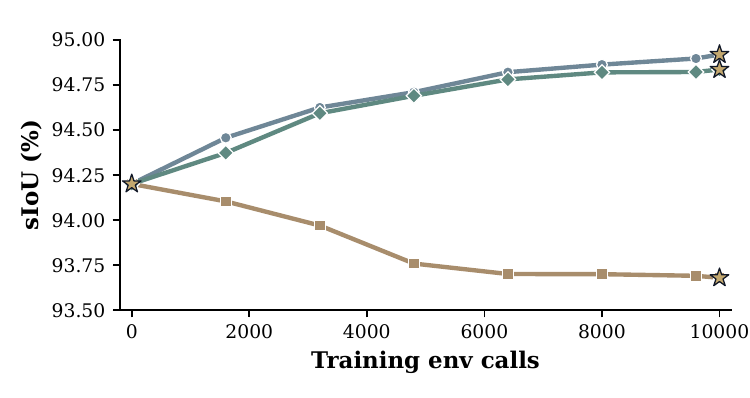}
    \end{minipage}\hfill
    \begin{minipage}[t]{0.48\textwidth}
        \centering
        \makebox[\linewidth][l]{\textbf{(b)}}\\[-0.25em]
        \includegraphics[width=\linewidth]{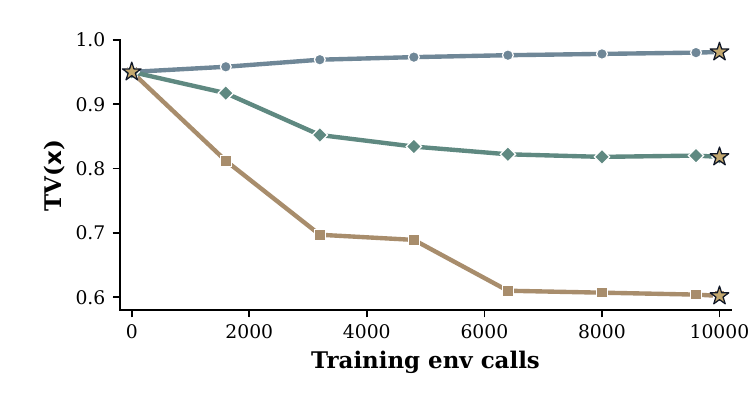}
    \end{minipage}\\[0.8em]
    \begin{minipage}{0.6\textwidth}
        \centering
        \makebox[\linewidth][l]{\textbf{(c)}}\\[-0.25em]
        \includegraphics[width=\linewidth]{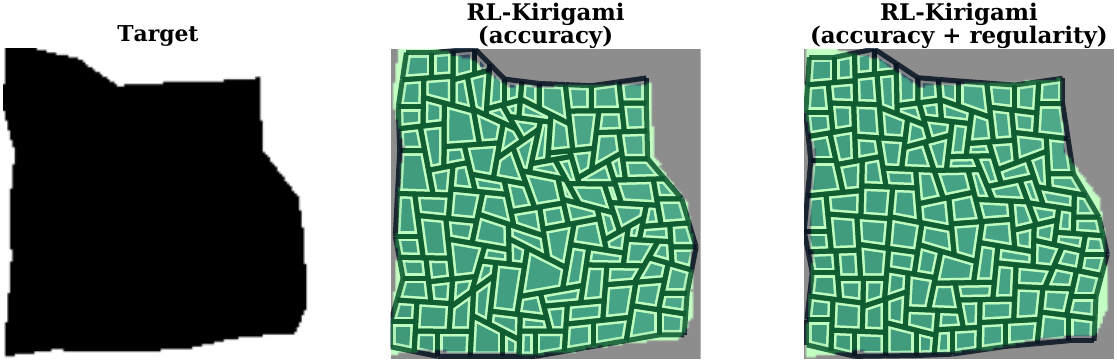}
    \end{minipage}
    \caption{RL training-budget sensitivity and selected regularity example. (a) \(\mathrm{sIoU}\) vs.\ training environment calls. (b) \(\mathrm{TV}(\mathbf{x})\) vs.\ training environment calls. Star markers show the values reported in Table~\ref{tab:rl_results_new}. (c) One selected target used to illustrate the effect of regularity preference in RL. Left: target silhouette. Middle: accuracy-only RL output. Right: accuracy plus regularity RL output. Lower \(\mathrm{TV}(\mathbf{x})\) indicates higher ratio-field regularity.}
    \label{fig:table3_sensitivity}
\end{figure}

Fig.~\ref{fig:table3_sensitivity}(a,b) shows \(\mathrm{sIoU}\) and \(\mathrm{TV}(\mathbf{x})\) over training environment calls for the three RL variants. The star markers denote the values reported in Table~\ref{tab:rl_results_new}. Accuracy-only GRPO increases \(\mathrm{sIoU}\) while keeping \(\mathrm{TV}(\mathbf{x})\) near the OT-CFM prior. Regularity-only GRPO lowers \(\mathrm{TV}(\mathbf{x})\) and reduces \(\mathrm{sIoU}\) as training continues. The combined reward increases \(\mathrm{sIoU}\) early and lowers \(\mathrm{TV}(\mathbf{x})\) over training.

Fig.~\ref{fig:table3_sensitivity}(c) shows one selected target for the reported RL settings. The accuracy-only run has \(\mathrm{TV}(\mathbf{x})=1.21\) and \(\mathrm{sIoU}=95.1\%\). The accuracy plus regularity run has \(\mathrm{TV}(\mathbf{x})=0.67\) and \(\mathrm{sIoU}=94.7\%\).

\subsection{Design for fabrication demonstration on laser-cut polyamide kirigami metamaterials}
\label{subsec:prototype}

\begin{figure}[pos=!t]
    \centering
    \includegraphics[width=\textwidth]{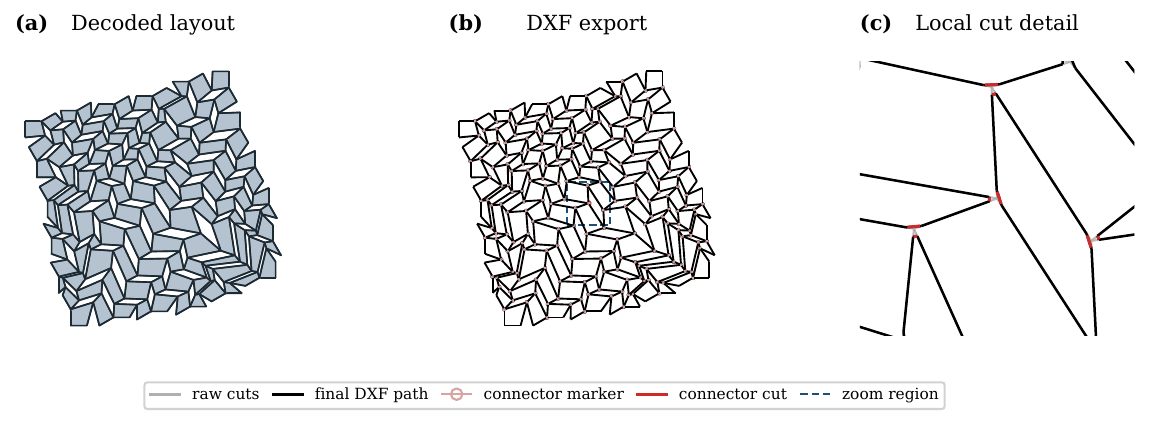}
    \caption{DXF export used for prototype fabrication. (a) A decoded layout from one generated ratio field \(\mathbf{x}\). (b) The corresponding cutter-ready DXF file, with the final cut path, connector markers, and the local connector cut highlighted in red. (c) Zoomed connector detail. Gray shows the raw local cuts, black shows the final DXF path, and red shows the connector cut kept around the marker.}
    \label{fig:supp_prototype_dxf_export}
\end{figure}

Three representative targets were selected for fabrication: a heart, a star, and a hexagon. For each shape, a design was first generated using the OT-CFM prior and the decoded flat layout from the ratio field \(\mathbf{x}\) was exported for laser cutting. Fig.~\ref{fig:supp_prototype_dxf_export} shows the decoded layout, the cutter-ready DXF file, and the local connector detail used in this export. Small connector markers are added, and the nearby cut paths are trimmed locally to preserve the connectors. The layout is then cut on a polyamide (PA, \(50~\mu\mathrm{m}\) thick) sheet as a sample polymeric material with a laser cutter (LPKF ProtoLaser U4), and the cut pieces are removed so the sheet can expand. The prototype is then ready for deployment at the target value of \(\varphi\) and overhead imaging.

\begin{figure}[pos=!htb]
    \centering
    \includegraphics[width=\textwidth]{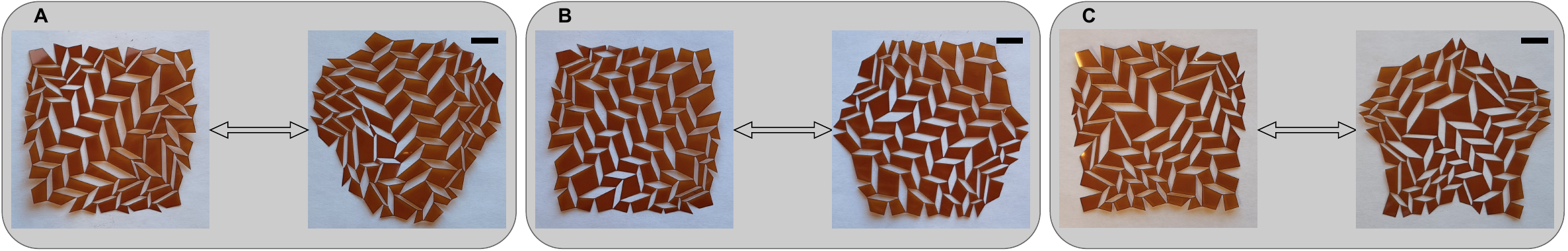}
    \caption{Rapid in situ prototyping on polyamide (PA) by laser cutting. (A) Heart, (B) hexagonal, and (C) star prototypes. In each panel, the left image shows the compact state and the right image shows the deployed state. Scale bars: \(1~\mathrm{cm}\).}
    \label{fig:prototype_pipeline}
\end{figure}

The reported outputs were the end-to-end prototype time together with rough stage-wise time estimates for generation, export, cutting, deployment, and imaging. Fig.~\ref{fig:prototype_pipeline} shows the fabricated polyamide prototypes side by side: (A) heart, (B) hexagonal, and (C) star. In each panel, the left image shows a compact state and the right image shows the deployed prototype state. In Supplementary Animation S1, the deployed motion of the fabricated star polyamide prototype is shown, and Supplementary Animation S2 shows the deployed motion of the fabricated hexagonal polyamide prototype. The full pipeline took \(8.0\pm 1.0\)~min per prototype, with less than \(10\)~s for pattern generation, less than \(10\)~s for export, about \(5\)~min for laser cutting, and about \(2\)~min for deployment and imaging. The total time was also affected by manual handling, but the measured end-to-end workflow still remained within minutes for each prototype.
In the reported examples, the laser-cut polyamide patterns in Fig.~\ref{fig:prototype_pipeline} transformed into the target shapes, whereas a similarly patterned aluminum sample exhibited connector failure during shape transformation, as shown in Fig.~\ref{fig:supp_metal_materials} (Appendix~\ref{supp:metal_trial}).

\section{Discussion and limitations}
\label{sec:discussion}

The first evaluation criterion in this study was whether RL-Kirigami can perform rapid inverse design for many targets under hard feasibility checks. Solver-based inverse design remains a strong and practical option, especially when only a small number of targets must be solved or when a per-target search loop is acceptable. Sec.~\ref{exp:fm_vs_solver} nevertheless shows that a data-driven method can be a strong alternative when many targets must be solved under the same geometry class, boundary conditions, and deployment setting. Fig.~\ref{fig:inverse_design_summary}(b)-(d) supports the same point by showing that this runtime advantage persists as grid size grows and that extra re-ranking is still much cheaper for the generator than repeating solver searches. In this sense, the large gap in forward-evaluation count matters more as the ratio-field dimension grows. The grid-size sweep is an empirical scaling result over the tested range, not a formal complexity claim. The main implication is that the present approach is most useful when many targets must be solved within a fixed task family, rather than as a replacement for all solver-based design.

The second evaluation criterion was to assess whether the base generator is reliable in the single-sample setting used by the full pipeline. Sec.~\ref{exp:gen_model_choice} justifies the choice of OT-CFM as the base generator for RL-Kirigami. Under the shared representation and geometric decoder, the highest overall accuracy and success rate were achieved among the tested conditional generators. The inference-step study in Fig.~\ref{fig:gen_steps_sensitivity}(a,b) also suggests that this conclusion is not tied to one narrow discretization choice. More importantly, OT-CFM remains competitive even at very low sampling-step counts, which is especially useful in a rapid inverse design pipeline. Diffusion reached a slightly higher IS, but its much longer sampling time made it less suitable for the present single-sample pipeline. This matters because the same target can admit multiple valid solutions, while the model still has to return a strong first sample. That is why the rest of RL-Kirigami was built on the OT-CFM prior rather than on the other generators.

The third evaluation criterion for the RL stage was to confirm it acts as preference alignment rather than replacement. As discussed in Sec.~\ref{exp:rl_acc_phys}, GRPO is most useful after pretraining, where it can shift the sampling distribution toward better task scores or higher ratio-field regularity, depending on the reward. Table~\ref{tab:rl_results_new} supports this interpretation. At the shared training budget, the accuracy-only reward mainly increases \(\mathrm{sIoU}\), the regularity-only reward lowers \(\mathrm{TV}(\mathbf{x})\) with a cost in \(\mathrm{sIoU}\), and the combined reward preserves most of the accuracy while also lowering \(\mathrm{TV}(\mathbf{x})\). The fact that different rewards move the model in different directions is itself useful, because it shows that RL is steering the generator toward the chosen preference rather than uniformly improving every metric. More generally, this means that the generator can be aligned directly to nondifferentiable task rewards and design preferences, such as the final accuracy score and the ratio-field regularity term used in this study, without requiring a differentiable surrogate. The budget study in Fig.~\ref{fig:table3_sensitivity}(a,b) supports the same reading by showing that the selected training budgets already capture the intended tradeoff. RL is therefore best understood here as a targeted refinement step on top of a pretrained OT-CFM generator, rather than as a replacement for it. Here, \(\mathrm{TV}(\mathbf{x})\) is used as a geometric regularity prior: lower local variation yields smoother transitions in cell morphology and more fabrication-friendly patterns, although it remains a geometric measure rather than a mechanical prediction.

A fourth evaluation criterion was whether the path from model output to fabrication can be kept short. Our laser cutting study in Sec.~\ref{subsec:prototype} supports this practical goal by showing that generated patterns can be exported to cutter-ready layouts and fabricated directly without manual pattern redesign. In the present setup, the decoded pattern can be converted deterministically and directly to a DXF file and immediately passed to the laser cutting workflow rather than being redrawn by hand or rebuilt in a separate design step. From a materials-design perspective, the fabrication result is not only a visual demonstration of shape matching; it also shows that the generated layouts can be translated into manufacturable cut paths while preserving local connectors that govern deployment. In the present setup, \(50~\mu\mathrm{m}\) polyamide accommodated the required panel rotation and connector compliance, whereas a similarly patterned aluminum trial failed at the connector regions (Fig.~\ref{fig:supp_metal_materials}). This contrast indicates that successful deployment depends jointly on cut topology, connector geometry, sheet thickness, and material compliance. The fabrication study is therefore best interpreted as proof-of-manufacturability and qualitative deployment validation, rather than full mechanical characterization.

These evaluation criteria also define the limits of the present study. We focused on compact reconfigurable parallelogram quad kirigami, fixed rectangular boundary anchors, and single-state silhouette matching at one fixed deployment parameter \(\varphi\). The method is planar, so spherical kirigami would require a different compatibility model and decoder. The geometry-based forward simulator does not model forces, material nonlinearity, contact along the full deployment path, or fabrication tolerances. The \(10\times 10\) grid and \(128\times 128\) masks also fix the design resolution. For this finite-sheet task, grid size is an empirical design-resolution choice rather than a universal representative length scale. The added \mbox{\(6\times 6\)} to \mbox{\(12\times 12\)} ablation in Appendix~\ref{supp:design_resolution_ablation} supports \mbox{\(10\times 10\)} as a practical setting for the present target family, while Fig.~\ref{fig:inverse_design_summary}(b) reports timing behavior up to \mbox{\(24\times 24\)}. Accordingly, the reported results provide evidence for fast inverse design within this constrained geometry class, simulator, and evaluator, rather than a full mechanics-aware inverse-design claim.

The OOD silhouette robustness test in Sec.~\ref{exp:ood_silhouette} further helps mark the applicability boundary of the current model, with representative failure cases in Fig.~\ref{fig:supp_failure_cases}. Lower scores on concave, literal, and targets with holes reflect both distribution shift and the geometric limits of the fixed compact parallelogram quad family under the chosen anchors and deployment setting. The direct solver shows a similar downward trend, with literal targets and targets with holes both scoring lowest. For OT-CFM, the with-holes group is lowest. For the solver, the literal group is slightly lower. These observations show a limitation of the tested family and setup, not a formal infeasibility proof for each target.

The present forward model is a geometry based deployment and rasterization model. It does not solve force balance and therefore does not predict stiffness, stress concentration, out of plane buckling, stress induced yielding, fracture, or fatigue. The broader framework can be generalized to such quantities when the evaluator is replaced or extended with a calibrated finite element model, a reduced order mechanical model, or an experimental measurement loop and those outputs are included in the reward. This paper demonstrates that idea only as a geometry-aware silhouette design and prototype manufacturability showcase, so mechanical prediction and validation remain future work.

An application direction for this kirigami family is the design of reconfigurable mechanisms and deployable tubular structures. Related work confirms that this class can behave as a reconfigurable rigid-deployable mechanism and can also be extended to three-dimensional tubular structures that morph between compact states~\citep{choi2021compact,dudte2023additive}. Another promising future direction is reconstructive-surgery design, especially meshed skin grafts, where mesh geometry affects achievable expansion~\citep{yu2025meshed}. While the present study does not model such settings directly, the same geometry class provides a compelling direction for future work.

\section{Conclusion}
\label{sec:conclusion}

RL-Kirigami combines an OT-CFM prior with GRPO fine-tuning for inverse kirigami design under hard feasibility checks and simulator-based objectives. Across the reported tests, the pretrained prior gave strong single-sample solutions with one forward evaluation, GRPO improved preference alignment to nondifferentiable task rewards and design preferences, and the generated patterns could be taken directly to rapid laser-cut fabrication. Overall, the reported results support a manufacturing-aware inverse design workflow for deployable kirigami metamaterials under hard geometric feasibility constraints.

\bibliographystyle{cas-model2-names}
\bibliography{refs}

\begin{thebibliography}{57}
\expandafter\ifx\csname natexlab\endcsname\relax\def\natexlab#1{#1}\fi
\providecommand{\url}[1]{\texttt{#1}}
\providecommand{\href}[2]{#2}
\providecommand{\path}[1]{#1}
\providecommand{\DOIprefix}{doi:}
\providecommand{\ArXivprefix}{arXiv:}
\providecommand{\URLprefix}{URL: }
\providecommand{\Pubmedprefix}{pmid:}
\providecommand{\doi}[1]{\href{http://dx.doi.org/#1}{\path{#1}}}
\providecommand{\Pubmed}[1]{\href{pmid:#1}{\path{#1}}}
\providecommand{\bibinfo}[2]{#2}
\ifx\xfnm\relax \def\xfnm[#1]{\unskip,\space#1}\fi
%Type = Article
\bibitem[{Albergo and Vanden-Eijnden(2023)}]{albergo2023building}
\bibinfo{author}{Albergo, M.}, \bibinfo{author}{Vanden-Eijnden, E.},
  \bibinfo{year}{2023}.
\newblock \bibinfo{title}{Building normalizing flows with stochastic
  interpolants}.
\newblock \bibinfo{journal}{International Conference on Learning
  Representations} .
%Type = Article
\bibitem[{Alderete et~al.(2022)Alderete, Pathak and
  Espinosa}]{alderete2022machine}
\bibinfo{author}{Alderete, N.A.}, \bibinfo{author}{Pathak, N.},
  \bibinfo{author}{Espinosa, H.D.}, \bibinfo{year}{2022}.
\newblock \bibinfo{title}{Machine learning assisted design of
  shape-programmable 3d kirigami metamaterials}.
\newblock \bibinfo{journal}{npj Computational Materials} \bibinfo{volume}{8},
  \bibinfo{pages}{191}.
%Type = Article
\bibitem[{Babaee et~al.(2021)Babaee, Shi, Abbasalizadeh, Tamang, Hess, Collins,
  Ishida, Lopes, Williams, Albaghdadi et~al.}]{babaee2021kirigami}
\bibinfo{author}{Babaee, S.}, \bibinfo{author}{Shi, Y.},
  \bibinfo{author}{Abbasalizadeh, S.}, \bibinfo{author}{Tamang, S.},
  \bibinfo{author}{Hess, K.}, \bibinfo{author}{Collins, J.E.},
  \bibinfo{author}{Ishida, K.}, \bibinfo{author}{Lopes, A.},
  \bibinfo{author}{Williams, M.}, \bibinfo{author}{Albaghdadi, M.}, et~al.,
  \bibinfo{year}{2021}.
\newblock \bibinfo{title}{Kirigami-inspired stents for sustained local delivery
  of therapeutics}.
\newblock \bibinfo{journal}{Nature Materials} \bibinfo{volume}{20},
  \bibinfo{pages}{1085--1092}.
%Type = Article
\bibitem[{Bastek and Kochmann(2023)}]{bastek2023inverse}
\bibinfo{author}{Bastek, J.H.}, \bibinfo{author}{Kochmann, D.M.},
  \bibinfo{year}{2023}.
\newblock \bibinfo{title}{Inverse design of nonlinear mechanical metamaterials
  via video denoising diffusion models}.
\newblock \bibinfo{journal}{Nature Machine Intelligence} \bibinfo{volume}{5},
  \bibinfo{pages}{1466--1475}.
%Type = Article
\bibitem[{Bertoldi et~al.(2017)Bertoldi, Vitelli, Christensen and
  Van~Hecke}]{bertoldi2017flexible}
\bibinfo{author}{Bertoldi, K.}, \bibinfo{author}{Vitelli, V.},
  \bibinfo{author}{Christensen, J.}, \bibinfo{author}{Van~Hecke, M.},
  \bibinfo{year}{2017}.
\newblock \bibinfo{title}{Flexible mechanical metamaterials}.
\newblock \bibinfo{journal}{Nature Reviews Materials} \bibinfo{volume}{2},
  \bibinfo{pages}{1--11}.
%Type = Article
\bibitem[{Blees et~al.(2015)Blees, Barnard, Rose, Roberts, McGill, Huang,
  Ruyack, Kevek, Kobrin, Muller et~al.}]{blees2015graphene}
\bibinfo{author}{Blees, M.K.}, \bibinfo{author}{Barnard, A.W.},
  \bibinfo{author}{Rose, P.A.}, \bibinfo{author}{Roberts, S.P.},
  \bibinfo{author}{McGill, K.L.}, \bibinfo{author}{Huang, P.Y.},
  \bibinfo{author}{Ruyack, A.R.}, \bibinfo{author}{Kevek, J.W.},
  \bibinfo{author}{Kobrin, B.}, \bibinfo{author}{Muller, D.A.}, et~al.,
  \bibinfo{year}{2015}.
\newblock \bibinfo{title}{Graphene kirigami}.
\newblock \bibinfo{journal}{Nature} \bibinfo{volume}{524},
  \bibinfo{pages}{204--207}.
%Type = Article
\bibitem[{Bliah et~al.(2025)Bliah, Hegde, Tan and
  Magdassi}]{bliah2025fabrication}
\bibinfo{author}{Bliah, O.}, \bibinfo{author}{Hegde, C.}, \bibinfo{author}{Tan,
  J.M.R.}, \bibinfo{author}{Magdassi, S.}, \bibinfo{year}{2025}.
\newblock \bibinfo{title}{Fabrication of soft robotics by additive
  manufacturing: from materials to applications}.
\newblock \bibinfo{journal}{Chemical Reviews} \bibinfo{volume}{125},
  \bibinfo{pages}{7275--7320}.
%Type = Article
\bibitem[{Boender et~al.(1982)Boender, Rinnooy~Kan, Timmer and
  Stougie}]{boender1982stochastic}
\bibinfo{author}{Boender, C.G.E.}, \bibinfo{author}{Rinnooy~Kan, A.},
  \bibinfo{author}{Timmer, G.}, \bibinfo{author}{Stougie, L.},
  \bibinfo{year}{1982}.
\newblock \bibinfo{title}{A stochastic method for global optimization}.
\newblock \bibinfo{journal}{Mathematical programming} \bibinfo{volume}{22},
  \bibinfo{pages}{125--140}.
%Type = Article
\bibitem[{Branyan et~al.(2022)Branyan, Rafsanjani, Bertoldi, Hatton and
  Meng{\"u}{\c{c}}}]{branyan2022curvilinear}
\bibinfo{author}{Branyan, C.}, \bibinfo{author}{Rafsanjani, A.},
  \bibinfo{author}{Bertoldi, K.}, \bibinfo{author}{Hatton, R.L.},
  \bibinfo{author}{Meng{\"u}{\c{c}}, Y.}, \bibinfo{year}{2022}.
\newblock \bibinfo{title}{Curvilinear kirigami skins let soft bending actuators
  slither faster}.
\newblock \bibinfo{journal}{Frontiers in Robotics and AI} \bibinfo{volume}{9},
  \bibinfo{pages}{872007}.
%Type = Article
\bibitem[{Brown et~al.(2023)Brown, Deshpande, Garland, Pradeep, Fadel, Pilla
  and Li}]{brown2023deep}
\bibinfo{author}{Brown, N.K.}, \bibinfo{author}{Deshpande, A.},
  \bibinfo{author}{Garland, A.}, \bibinfo{author}{Pradeep, S.A.},
  \bibinfo{author}{Fadel, G.}, \bibinfo{author}{Pilla, S.},
  \bibinfo{author}{Li, G.}, \bibinfo{year}{2023}.
\newblock \bibinfo{title}{Deep reinforcement learning for the design of
  mechanical metamaterials with tunable deformation and hysteretic
  characteristics}.
\newblock \bibinfo{journal}{Materials \& Design} \bibinfo{volume}{235},
  \bibinfo{pages}{112428}.
%Type = Article
\bibitem[{Brown et~al.(2022)Brown, Garland, Fadel and Li}]{brown2022deep}
\bibinfo{author}{Brown, N.K.}, \bibinfo{author}{Garland, A.P.},
  \bibinfo{author}{Fadel, G.M.}, \bibinfo{author}{Li, G.},
  \bibinfo{year}{2022}.
\newblock \bibinfo{title}{Deep reinforcement learning for engineering design
  through topology optimization of elementally discretized design domains}.
\newblock \bibinfo{journal}{Materials \& Design} \bibinfo{volume}{218},
  \bibinfo{pages}{110672}.
%Type = Article
\bibitem[{Brzin et~al.(2025)Brzin, Jawed and Brojan}]{brzin2025generative}
\bibinfo{author}{Brzin, T.}, \bibinfo{author}{Jawed, M.K.},
  \bibinfo{author}{Brojan, M.}, \bibinfo{year}{2025}.
\newblock \bibinfo{title}{Generative adversarial network-based inverse design
  of self-deploying soft kirigami composites for targeted shape
  transformation}.
\newblock \bibinfo{journal}{Engineering Applications of Artificial
  Intelligence} \bibinfo{volume}{149}, \bibinfo{pages}{110417}.
%Type = Article
\bibitem[{Cerniauskas et~al.(2024)Cerniauskas, Sadia and
  Alam}]{cerniauskas2024machine}
\bibinfo{author}{Cerniauskas, G.}, \bibinfo{author}{Sadia, H.},
  \bibinfo{author}{Alam, P.}, \bibinfo{year}{2024}.
\newblock \bibinfo{title}{Machine intelligence in metamaterials design: a
  review}.
\newblock \bibinfo{journal}{Oxford Open Materials Science} \bibinfo{volume}{4},
  \bibinfo{pages}{itae001}.
%Type = Article
\bibitem[{Chen et~al.(2025)Chen, Sun, Lee, Portela and
  Chen}]{chen2025generative}
\bibinfo{author}{Chen, W.}, \bibinfo{author}{Sun, R.}, \bibinfo{author}{Lee,
  D.}, \bibinfo{author}{Portela, C.M.}, \bibinfo{author}{Chen, W.},
  \bibinfo{year}{2025}.
\newblock \bibinfo{title}{Generative inverse design of metamaterials with
  functional responses by interpretable learning}.
\newblock \bibinfo{journal}{Advanced Intelligent Systems} \bibinfo{volume}{7},
  \bibinfo{pages}{2400611}.
%Type = Article
\bibitem[{Choi et~al.(2019)Choi, Dudte and Mahadevan}]{choi2019programming}
\bibinfo{author}{Choi, G.P.}, \bibinfo{author}{Dudte, L.H.},
  \bibinfo{author}{Mahadevan, L.}, \bibinfo{year}{2019}.
\newblock \bibinfo{title}{Programming shape using kirigami tessellations}.
\newblock \bibinfo{journal}{Nature materials} \bibinfo{volume}{18},
  \bibinfo{pages}{999--1004}.
%Type = Article
\bibitem[{Choi et~al.(2021)Choi, Dudte and Mahadevan}]{choi2021compact}
\bibinfo{author}{Choi, G.P.}, \bibinfo{author}{Dudte, L.H.},
  \bibinfo{author}{Mahadevan, L.}, \bibinfo{year}{2021}.
\newblock \bibinfo{title}{Compact reconfigurable kirigami}.
\newblock \bibinfo{journal}{Physical Review Research} \bibinfo{volume}{3},
  \bibinfo{pages}{043030}.
%Type = Article
\bibitem[{Dang et~al.(2021)Dang, Feng, Duan and Wang}]{dang2021deployable}
\bibinfo{author}{Dang, X.}, \bibinfo{author}{Feng, F.}, \bibinfo{author}{Duan,
  H.}, \bibinfo{author}{Wang, J.}, \bibinfo{year}{2021}.
\newblock \bibinfo{title}{Theorem for the design of deployable kirigami
  tessellations with different topologies}.
\newblock \bibinfo{journal}{Physical Review E} \bibinfo{volume}{104},
  \bibinfo{pages}{055006}.
%Type = Article
\bibitem[{Dang et~al.(2022)Dang, Feng, Duan and Wang}]{dang2022spherical}
\bibinfo{author}{Dang, X.}, \bibinfo{author}{Feng, F.}, \bibinfo{author}{Duan,
  H.}, \bibinfo{author}{Wang, J.}, \bibinfo{year}{2022}.
\newblock \bibinfo{title}{Theorem on the compatibility of spherical kirigami
  tessellations}.
\newblock \bibinfo{journal}{Physical Review Letters} \bibinfo{volume}{128},
  \bibinfo{pages}{035501}.
%Type = Article
\bibitem[{Dudte et~al.(2023)Dudte, Choi, Becker and
  Mahadevan}]{dudte2023additive}
\bibinfo{author}{Dudte, L.H.}, \bibinfo{author}{Choi, G.P.},
  \bibinfo{author}{Becker, K.P.}, \bibinfo{author}{Mahadevan, L.},
  \bibinfo{year}{2023}.
\newblock \bibinfo{title}{An additive framework for kirigami design}.
\newblock \bibinfo{journal}{Nature Computational Science} \bibinfo{volume}{3},
  \bibinfo{pages}{443--454}.
%Type = Article
\bibitem[{Felsch and Slesarenko(2024)}]{felsch2024generative}
\bibinfo{author}{Felsch, G.}, \bibinfo{author}{Slesarenko, V.},
  \bibinfo{year}{2024}.
\newblock \bibinfo{title}{Generative models struggle with kirigami
  metamaterials}.
\newblock \bibinfo{journal}{Scientific Reports} \bibinfo{volume}{14},
  \bibinfo{pages}{19397}.
%Type = Article
\bibitem[{Gower(1975)}]{gower1975generalized}
\bibinfo{author}{Gower, J.C.}, \bibinfo{year}{1975}.
\newblock \bibinfo{title}{Generalized procrustes analysis}.
\newblock \bibinfo{journal}{Psychometrika} \bibinfo{volume}{40},
  \bibinfo{pages}{33--51}.
%Type = Article
\bibitem[{Ha et~al.(2023)Ha, Yao, Xu, Liu, Liu, Elkins, Kile, Deshpande, Kong,
  Bauchy et~al.}]{ha2023rapid}
\bibinfo{author}{Ha, C.S.}, \bibinfo{author}{Yao, D.}, \bibinfo{author}{Xu,
  Z.}, \bibinfo{author}{Liu, C.}, \bibinfo{author}{Liu, H.},
  \bibinfo{author}{Elkins, D.}, \bibinfo{author}{Kile, M.},
  \bibinfo{author}{Deshpande, V.}, \bibinfo{author}{Kong, Z.},
  \bibinfo{author}{Bauchy, M.}, et~al., \bibinfo{year}{2023}.
\newblock \bibinfo{title}{Rapid inverse design of metamaterials based on
  prescribed mechanical behavior through machine learning}.
\newblock \bibinfo{journal}{Nature Communications} \bibinfo{volume}{14},
  \bibinfo{pages}{5765}.
%Type = Article
\bibitem[{Hansen and Ostermeier(2001)}]{hansen2001completely}
\bibinfo{author}{Hansen, N.}, \bibinfo{author}{Ostermeier, A.},
  \bibinfo{year}{2001}.
\newblock \bibinfo{title}{Completely derandomized self-adaptation in evolution
  strategies}.
\newblock \bibinfo{journal}{Evolutionary computation} \bibinfo{volume}{9},
  \bibinfo{pages}{159--195}.
%Type = Article
\bibitem[{Jolliffe and Cadima(2016)}]{jolliffe2016principal}
\bibinfo{author}{Jolliffe, I.T.}, \bibinfo{author}{Cadima, J.},
  \bibinfo{year}{2016}.
\newblock \bibinfo{title}{Principal component analysis: a review and recent
  developments}.
\newblock \bibinfo{journal}{Philosophical transactions of the royal society A:
  Mathematical, Physical and Engineering Sciences} \bibinfo{volume}{374},
  \bibinfo{pages}{20150202}.
%Type = Article
\bibitem[{Khosravi and Li(2022)}]{khosravi2022tunable}
\bibinfo{author}{Khosravi, H.}, \bibinfo{author}{Li, S.}, \bibinfo{year}{2022}.
\newblock \bibinfo{title}{Tunable wave-propagation band gap via stretching
  kirigami sheets}.
\newblock \bibinfo{journal}{Physical Review Applied} \bibinfo{volume}{17},
  \bibinfo{pages}{064054}.
%Type = Article
\bibitem[{Kim and Lee(2018)}]{kim2018shape}
\bibinfo{author}{Kim, T.}, \bibinfo{author}{Lee, Y.G.}, \bibinfo{year}{2018}.
\newblock \bibinfo{title}{Shape transformable bifurcated stents}.
\newblock \bibinfo{journal}{Scientific reports} \bibinfo{volume}{8},
  \bibinfo{pages}{13911}.
%Type = Article
\bibitem[{Lamoureux et~al.(2015)Lamoureux, Lee, Shlian, Forrest and
  Shtein}]{lamoureux2015dynamic}
\bibinfo{author}{Lamoureux, A.}, \bibinfo{author}{Lee, K.},
  \bibinfo{author}{Shlian, M.}, \bibinfo{author}{Forrest, S.R.},
  \bibinfo{author}{Shtein, M.}, \bibinfo{year}{2015}.
\newblock \bibinfo{title}{Dynamic kirigami structures for integrated solar
  tracking}.
\newblock \bibinfo{journal}{Nature communications} \bibinfo{volume}{6},
  \bibinfo{pages}{8092}.
%Type = Article
\bibitem[{Lipman et~al.(2023)Lipman, Chen, Ben-Hamu, Nickel and
  Le}]{lipman2023flow}
\bibinfo{author}{Lipman, Y.}, \bibinfo{author}{Chen, R.T.},
  \bibinfo{author}{Ben-Hamu, H.}, \bibinfo{author}{Nickel, M.},
  \bibinfo{author}{Le, M.}, \bibinfo{year}{2023}.
\newblock \bibinfo{title}{Flow matching for generative modeling}.
\newblock \bibinfo{journal}{International Conference on Learning
  Representations} .
%Type = Article
\bibitem[{Liu et~al.(2023)Liu, Gong and Liu}]{liu2023flow}
\bibinfo{author}{Liu, X.}, \bibinfo{author}{Gong, C.}, \bibinfo{author}{Liu,
  Q.}, \bibinfo{year}{2023}.
\newblock \bibinfo{title}{Flow straight and fast: Learning to generate and
  transfer data with rectified flow}.
\newblock \bibinfo{journal}{International Conference on Learning
  Representations} .
%Type = Article
\bibitem[{Mureau et~al.(2005)Mureau, Posch, Meeuwis and
  Hofer}]{mureau2005anterolateral}
\bibinfo{author}{Mureau, M.A.}, \bibinfo{author}{Posch, N.A.},
  \bibinfo{author}{Meeuwis, C.A.}, \bibinfo{author}{Hofer, S.O.},
  \bibinfo{year}{2005}.
\newblock \bibinfo{title}{Anterolateral thigh flap reconstruction of large
  external facial skin defects: a follow-up study on functional and aesthetic
  recipient-and donor-site outcome}.
\newblock \bibinfo{journal}{Plastic and reconstructive surgery}
  \bibinfo{volume}{115}, \bibinfo{pages}{1077--1086}.
%Type = Inproceedings
\bibitem[{Park et~al.(2019)Park, Liu, Wang and Zhu}]{park2019semantic}
\bibinfo{author}{Park, T.}, \bibinfo{author}{Liu, M.Y.}, \bibinfo{author}{Wang,
  T.C.}, \bibinfo{author}{Zhu, J.Y.}, \bibinfo{year}{2019}.
\newblock \bibinfo{title}{Semantic image synthesis with spatially-adaptive
  normalization}, in: \bibinfo{booktitle}{Proceedings of the IEEE/CVF
  conference on computer vision and pattern recognition}, pp.
  \bibinfo{pages}{2337--2346}.
%Type = Article
\bibitem[{Perez and Behdinan(2007)}]{perez2007particle}
\bibinfo{author}{Perez, R.E.}, \bibinfo{author}{Behdinan, K.},
  \bibinfo{year}{2007}.
\newblock \bibinfo{title}{Particle swarm approach for structural design
  optimization}.
\newblock \bibinfo{journal}{Computers \& Structures} \bibinfo{volume}{85},
  \bibinfo{pages}{1579--1588}.
%Type = Article
\bibitem[{Powell(1964)}]{powell1964efficient}
\bibinfo{author}{Powell, M.J.}, \bibinfo{year}{1964}.
\newblock \bibinfo{title}{An efficient method for finding the minimum of a
  function of several variables without calculating derivatives}.
\newblock \bibinfo{journal}{The computer journal} \bibinfo{volume}{7},
  \bibinfo{pages}{155--162}.
%Type = Article
\bibitem[{Qiao et~al.(2025)Qiao, Chen, Chen, Zhou, Jiang, Wang, Tian and
  Pasini}]{qiao2025inverse}
\bibinfo{author}{Qiao, C.}, \bibinfo{author}{Chen, S.}, \bibinfo{author}{Chen,
  Y.}, \bibinfo{author}{Zhou, Z.}, \bibinfo{author}{Jiang, W.},
  \bibinfo{author}{Wang, Q.}, \bibinfo{author}{Tian, X.},
  \bibinfo{author}{Pasini, D.}, \bibinfo{year}{2025}.
\newblock \bibinfo{title}{Inverse design of kirigami through shape programming
  of rotating units}.
\newblock \bibinfo{journal}{Physical Review Letters} \bibinfo{volume}{134},
  \bibinfo{pages}{176103}.
%Type = Article
\bibitem[{Rafsanjani et~al.(2018)Rafsanjani, Zhang, Liu, Rubinstein and
  Bertoldi}]{rafsanjani2018kirigami}
\bibinfo{author}{Rafsanjani, A.}, \bibinfo{author}{Zhang, Y.},
  \bibinfo{author}{Liu, B.}, \bibinfo{author}{Rubinstein, S.M.},
  \bibinfo{author}{Bertoldi, K.}, \bibinfo{year}{2018}.
\newblock \bibinfo{title}{Kirigami skins make a simple soft actuator crawl}.
\newblock \bibinfo{journal}{Science Robotics} \bibinfo{volume}{3},
  \bibinfo{pages}{eaar7555}.
%Type = Article
\bibitem[{Rosafalco et~al.(2023)Rosafalco, De~Ponti, Iorio, Craster, Ardito and
  Corigliano}]{rosafalco2023reinforcement}
\bibinfo{author}{Rosafalco, L.}, \bibinfo{author}{De~Ponti, J.M.},
  \bibinfo{author}{Iorio, L.}, \bibinfo{author}{Craster, R.V.},
  \bibinfo{author}{Ardito, R.}, \bibinfo{author}{Corigliano, A.},
  \bibinfo{year}{2023}.
\newblock \bibinfo{title}{Reinforcement learning optimisation for graded
  metamaterial design using a physical-based constraint on the state
  representation and action space}.
\newblock \bibinfo{journal}{Scientific Reports} \bibinfo{volume}{13},
  \bibinfo{pages}{21836}.
%Type = Article
\bibitem[{Salimans et~al.(2016)Salimans, Goodfellow, Zaremba, Cheung, Radford
  and Chen}]{salimans2016improved}
\bibinfo{author}{Salimans, T.}, \bibinfo{author}{Goodfellow, I.},
  \bibinfo{author}{Zaremba, W.}, \bibinfo{author}{Cheung, V.},
  \bibinfo{author}{Radford, A.}, \bibinfo{author}{Chen, X.},
  \bibinfo{year}{2016}.
\newblock \bibinfo{title}{Improved techniques for training gans}.
\newblock \bibinfo{journal}{Advances in neural information processing systems}
  \bibinfo{volume}{29}.
%Type = Article
\bibitem[{Shah et~al.(2021)Shah, Zhuo, Lai, De~La Rosa-Moreno, Amirkulova and
  Gerstoft}]{shah2021reinforcement}
\bibinfo{author}{Shah, T.}, \bibinfo{author}{Zhuo, L.}, \bibinfo{author}{Lai,
  P.}, \bibinfo{author}{De~La Rosa-Moreno, A.}, \bibinfo{author}{Amirkulova,
  F.}, \bibinfo{author}{Gerstoft, P.}, \bibinfo{year}{2021}.
\newblock \bibinfo{title}{Reinforcement learning applied to metamaterial
  design}.
\newblock \bibinfo{journal}{The Journal of the Acoustical Society of America}
  \bibinfo{volume}{150}, \bibinfo{pages}{321--338}.
%Type = Article
\bibitem[{Shao et~al.(2024)Shao, Wang, Zhu, Xu, Song, Bi, Zhang, Zhang, Li, Wu
  et~al.}]{shao2024deepseekmath}
\bibinfo{author}{Shao, Z.}, \bibinfo{author}{Wang, P.}, \bibinfo{author}{Zhu,
  Q.}, \bibinfo{author}{Xu, R.}, \bibinfo{author}{Song, J.},
  \bibinfo{author}{Bi, X.}, \bibinfo{author}{Zhang, H.},
  \bibinfo{author}{Zhang, M.}, \bibinfo{author}{Li, Y.}, \bibinfo{author}{Wu,
  Y.}, et~al., \bibinfo{year}{2024}.
\newblock \bibinfo{title}{Deepseekmath: Pushing the limits of mathematical
  reasoning in open language models}.
\newblock \bibinfo{journal}{arXiv preprint arXiv:2402.03300} .
%Type = Article
\bibitem[{Sobol(1967)}]{sobol1967distribution}
\bibinfo{author}{Sobol, I.M.}, \bibinfo{year}{1967}.
\newblock \bibinfo{title}{Distribution of points in a cube and approximate
  evaluation of integrals}.
\newblock \bibinfo{journal}{USSR Computational mathematics and mathematical
  physics} \bibinfo{volume}{7}, \bibinfo{pages}{86--112}.
%Type = Article
\bibitem[{Song et~al.(2021)Song, Meng and Ermon}]{song2021ddim}
\bibinfo{author}{Song, J.}, \bibinfo{author}{Meng, C.}, \bibinfo{author}{Ermon,
  S.}, \bibinfo{year}{2021}.
\newblock \bibinfo{title}{Denoising diffusion implicit models}.
\newblock \bibinfo{journal}{International Conference on Learning
  Representations} .
%Type = Article
\bibitem[{Sun et~al.(2018)Sun, Zhang, Yang, Zhang, Farrow, Scarpa and
  Rossiter}]{sun2018kirigami}
\bibinfo{author}{Sun, R.}, \bibinfo{author}{Zhang, B.}, \bibinfo{author}{Yang,
  L.}, \bibinfo{author}{Zhang, W.}, \bibinfo{author}{Farrow, I.},
  \bibinfo{author}{Scarpa, F.}, \bibinfo{author}{Rossiter, J.},
  \bibinfo{year}{2018}.
\newblock \bibinfo{title}{Kirigami stretchable strain sensors with enhanced
  piezoelectricity induced by topological electrodes}.
\newblock \bibinfo{journal}{Applied Physics Letters} \bibinfo{volume}{112}.
%Type = Article
\bibitem[{Tirado et~al.(2025)Tirado, Parvaresh, Seyido{\u{g}}lu, Bedford,
  J{\o}rgensen and Rafsanjani}]{tirado2025multimodal}
\bibinfo{author}{Tirado, J.}, \bibinfo{author}{Parvaresh, A.},
  \bibinfo{author}{Seyido{\u{g}}lu, B.}, \bibinfo{author}{Bedford, D.A.},
  \bibinfo{author}{J{\o}rgensen, J.}, \bibinfo{author}{Rafsanjani, A.},
  \bibinfo{year}{2025}.
\newblock \bibinfo{title}{Multimodal limbless crawling soft robot with a
  kirigami skin}.
\newblock \bibinfo{journal}{Cyborg and Bionic Systems} \bibinfo{volume}{6},
  \bibinfo{pages}{0301}.
%Type = Article
\bibitem[{Wang et~al.(2022)Wang, Zhang, Li and You}]{wang2022kirigami}
\bibinfo{author}{Wang, C.}, \bibinfo{author}{Zhang, D.}, \bibinfo{author}{Li,
  J.}, \bibinfo{author}{You, Z.}, \bibinfo{year}{2022}.
\newblock \bibinfo{title}{Kirigami-inspired thick-panel deployable structures}.
\newblock \bibinfo{journal}{International Journal of Solids and Structures}
  \bibinfo{volume}{251}, \bibinfo{pages}{111752}.
%Type = Article
\bibitem[{Wei et~al.(2024)Wei, Li, Chen, Hua, Shen and
  Jiang}]{wei2024revolutionizing}
\bibinfo{author}{Wei, R.}, \bibinfo{author}{Li, H.}, \bibinfo{author}{Chen,
  Z.}, \bibinfo{author}{Hua, Q.}, \bibinfo{author}{Shen, G.},
  \bibinfo{author}{Jiang, K.}, \bibinfo{year}{2024}.
\newblock \bibinfo{title}{Revolutionizing wearable technology: advanced
  fabrication techniques for body-conformable electronics}.
\newblock \bibinfo{journal}{npj Flexible Electronics} \bibinfo{volume}{8},
  \bibinfo{pages}{83}.
%Type = Article
\bibitem[{Won et~al.(2019)Won, Park, Lee, Ha, Han, Choi, Lee, Hong, Cho and
  Ko}]{won2019stretchable}
\bibinfo{author}{Won, P.}, \bibinfo{author}{Park, J.J.}, \bibinfo{author}{Lee,
  T.}, \bibinfo{author}{Ha, I.}, \bibinfo{author}{Han, S.},
  \bibinfo{author}{Choi, M.}, \bibinfo{author}{Lee, J.}, \bibinfo{author}{Hong,
  S.}, \bibinfo{author}{Cho, K.J.}, \bibinfo{author}{Ko, S.H.},
  \bibinfo{year}{2019}.
\newblock \bibinfo{title}{Stretchable and transparent kirigami conductor of
  nanowire percolation network for electronic skin applications}.
\newblock \bibinfo{journal}{Nano letters} \bibinfo{volume}{19},
  \bibinfo{pages}{6087--6096}.
%Type = Article
\bibitem[{Xiang et~al.(2024)Xiang, Hou, Chen, Pidaparti, Song, Tang and
  Wang}]{xiang2024gan}
\bibinfo{author}{Xiang, Y.}, \bibinfo{author}{Hou, J.}, \bibinfo{author}{Chen,
  X.}, \bibinfo{author}{Pidaparti, R.}, \bibinfo{author}{Song, K.},
  \bibinfo{author}{Tang, K.}, \bibinfo{author}{Wang, X.}, \bibinfo{year}{2024}.
\newblock \bibinfo{title}{A gan-based stepwise full-field mechanical prediction
  model for architected metamaterials}.
\newblock \bibinfo{journal}{International Journal of Mechanical Sciences}
  \bibinfo{volume}{284}, \bibinfo{pages}{109771}.
%Type = Article
\bibitem[{Xue et~al.(2017)Xue, Li, Du, Zhang, Zhu, Sun and
  Guo}]{xue2017kirigami}
\bibinfo{author}{Xue, R.}, \bibinfo{author}{Li, R.}, \bibinfo{author}{Du, Z.},
  \bibinfo{author}{Zhang, W.}, \bibinfo{author}{Zhu, Y.}, \bibinfo{author}{Sun,
  Z.}, \bibinfo{author}{Guo, X.}, \bibinfo{year}{2017}.
\newblock \bibinfo{title}{Kirigami pattern design of mechanically driven
  formation of complex 3d structures through topology optimization}.
\newblock \bibinfo{journal}{Extreme Mechanics Letters} \bibinfo{volume}{15},
  \bibinfo{pages}{139--144}.
%Type = Article
\bibitem[{Yang et~al.(2018)Yang, Dias and Holmes}]{yang2018multistable}
\bibinfo{author}{Yang, Y.}, \bibinfo{author}{Dias, M.A.},
  \bibinfo{author}{Holmes, D.P.}, \bibinfo{year}{2018}.
\newblock \bibinfo{title}{Multistable kirigami for tunable architected
  materials}.
\newblock \bibinfo{journal}{Physical Review Materials} \bibinfo{volume}{2},
  \bibinfo{pages}{110601}.
%Type = Article
\bibitem[{Yang et~al.(2026)Yang, Wang, Zhai, Chen, Wu, Zhao, Chen, Liu and
  Fu}]{yang2026guided}
\bibinfo{author}{Yang, Y.}, \bibinfo{author}{Wang, L.}, \bibinfo{author}{Zhai,
  X.}, \bibinfo{author}{Chen, K.}, \bibinfo{author}{Wu, W.},
  \bibinfo{author}{Zhao, Y.}, \bibinfo{author}{Chen, F.}, \bibinfo{author}{Liu,
  L.}, \bibinfo{author}{Fu, X.M.}, \bibinfo{year}{2026}.
\newblock \bibinfo{title}{Guided diffusion for fast inverse design of
  voxel-based mechanical metamaterials}.
\newblock \bibinfo{journal}{Smart Materials in Manufacturing}
  \bibinfo{volume}{4}, \bibinfo{pages}{100129}.
%Type = Inproceedings
\bibitem[{Yazdani et~al.(2025)Yazdani, Medghalchi, Ashrafian, Hacihaliloglu and
  Shahriari}]{yazdani2025flow}
\bibinfo{author}{Yazdani, M.}, \bibinfo{author}{Medghalchi, Y.},
  \bibinfo{author}{Ashrafian, P.}, \bibinfo{author}{Hacihaliloglu, I.},
  \bibinfo{author}{Shahriari, D.}, \bibinfo{year}{2025}.
\newblock \bibinfo{title}{Flow matching for medical image synthesis: Bridging
  the gap between speed and quality}, in: \bibinfo{booktitle}{International
  Conference on Medical Image Computing and Computer-Assisted Intervention},
  \bibinfo{organization}{Springer}. pp. \bibinfo{pages}{216--226}.
%Type = Article
\bibitem[{Ying et~al.(2025)Ying, Fernando and Dias}]{ying2025inverse}
\bibinfo{author}{Ying, X.}, \bibinfo{author}{Fernando, D.},
  \bibinfo{author}{Dias, M.A.}, \bibinfo{year}{2025}.
\newblock \bibinfo{title}{Inverse design of programmable shape-morphing
  kirigami structures}.
\newblock \bibinfo{journal}{International Journal of Mechanical Sciences}
  \bibinfo{volume}{286}, \bibinfo{pages}{109840}.
%Type = Article
\bibitem[{Yu et~al.(2025)Yu, Jafari, Mujahid, Garcia, Shah, Sinha, Huang,
  Shakiba, Hong, Cheraghali et~al.}]{yu2025meshed}
\bibinfo{author}{Yu, H.}, \bibinfo{author}{Jafari, M.},
  \bibinfo{author}{Mujahid, A.}, \bibinfo{author}{Garcia, C.F.},
  \bibinfo{author}{Shah, J.}, \bibinfo{author}{Sinha, R.},
  \bibinfo{author}{Huang, Y.}, \bibinfo{author}{Shakiba, D.},
  \bibinfo{author}{Hong, Y.}, \bibinfo{author}{Cheraghali, D.}, et~al.,
  \bibinfo{year}{2025}.
\newblock \bibinfo{title}{Expansion limits of meshed split-thickness skin
  grafts}.
\newblock \bibinfo{journal}{Acta biomaterialia} \bibinfo{volume}{191},
  \bibinfo{pages}{325--335}.
%Type = Article
\bibitem[{Zhai et~al.(2021)Zhai, Wu and Jiang}]{zhai2021mechanical}
\bibinfo{author}{Zhai, Z.}, \bibinfo{author}{Wu, L.}, \bibinfo{author}{Jiang,
  H.}, \bibinfo{year}{2021}.
\newblock \bibinfo{title}{Mechanical metamaterials based on origami and
  kirigami}.
\newblock \bibinfo{journal}{Applied Physics Reviews} \bibinfo{volume}{8}.
%Type = Inproceedings
\bibitem[{Zhang et~al.(2023)Zhang, Rao and Agrawala}]{zhang2023controlnet}
\bibinfo{author}{Zhang, L.}, \bibinfo{author}{Rao, A.},
  \bibinfo{author}{Agrawala, M.}, \bibinfo{year}{2023}.
\newblock \bibinfo{title}{Adding conditional control to text-to-image diffusion
  models}, in: \bibinfo{booktitle}{Proceedings of the IEEE/CVF international
  conference on computer vision}, pp. \bibinfo{pages}{3836--3847}.
%Type = Article
\bibitem[{Zheng et~al.(2023)Zheng, Zhang, Chen and Watanabe}]{zheng2023deep}
\bibinfo{author}{Zheng, X.}, \bibinfo{author}{Zhang, X.},
  \bibinfo{author}{Chen, T.T.}, \bibinfo{author}{Watanabe, I.},
  \bibinfo{year}{2023}.
\newblock \bibinfo{title}{Deep learning in mechanical metamaterials: from
  prediction and generation to inverse design}.
\newblock \bibinfo{journal}{Advanced Materials} \bibinfo{volume}{35},
  \bibinfo{pages}{2302530}.
%Type = Article
\bibitem[{Zheng et~al.(2022)Zheng, Niloy, Celli, Tobasco and
  Plucinsky}]{zheng2022continuum}
\bibinfo{author}{Zheng, Y.}, \bibinfo{author}{Niloy, I.},
  \bibinfo{author}{Celli, P.}, \bibinfo{author}{Tobasco, I.},
  \bibinfo{author}{Plucinsky, P.}, \bibinfo{year}{2022}.
\newblock \bibinfo{title}{Continuum field theory for the deformations of planar
  kirigami}.
\newblock \bibinfo{journal}{Physical review letters} \bibinfo{volume}{128},
  \bibinfo{pages}{208003}.

\end{thebibliography}
\clearpage
\appendix
\numberwithin{equation}{section}
\numberwithin{figure}{section}
\numberwithin{table}{section}
\renewcommand{\theHsection}{appendix.\thesection}
\renewcommand{\theHsubsection}{appendix.\thesubsection}
\renewcommand{\theHfigure}{appendix.\thesection.\arabic{figure}}
\renewcommand{\theHtable}{appendix.\thesection.\arabic{table}}
\renewcommand{\theHequation}{appendix.\thesection.\arabic{equation}}

\section{Materials and Methods}
\label{supp:materials_methods}

\subsection{Hardware}
\label{supp:hardware}
All training, inference, and solver experiments were run on a workstation with an AMD Ryzen Threadripper PRO 7965WX CPU, \(125~\mathrm{GB}\) system memory, and two NVIDIA GeForce RTX 4090 GPUs with \(24~\mathrm{GB}\) memory each. The reported wall-clock times in Sec.~\ref{exp:fm_vs_solver}, Table~\ref{tab:gen_models}, and Sec.~\ref{subsec:prototype} are hardware-dependent and correspond to this setup. Prototype fabrication used the same LPKF ProtoLaser U4 setup reported in Sec.~\ref{subsec:prototype}.

\subsection{Additional material trial}
\label{supp:metal_trial}
\begin{figure}[pos=t]
    \centering
    \includegraphics[width=0.45\textwidth]{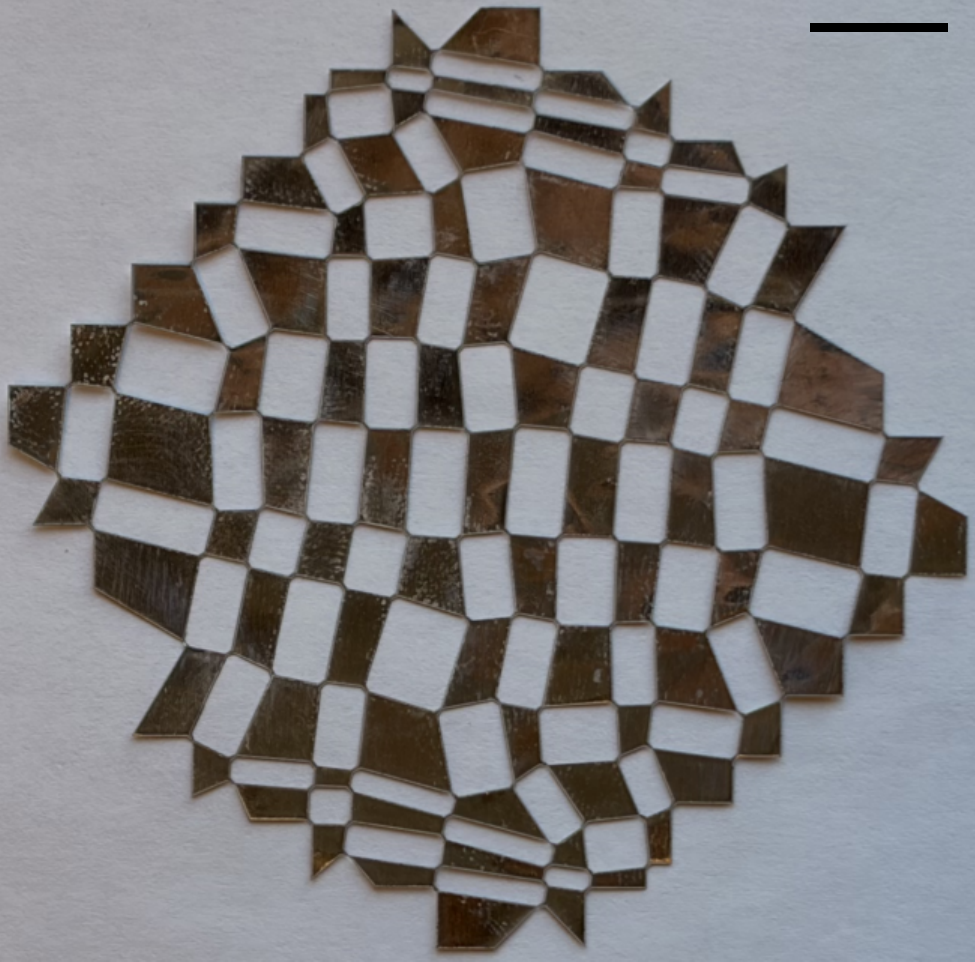}
    \caption{Additional material trial with aluminum. The generated cut layout could be laser cut, but the connector regions were not compliant enough to allow the panel rotation and deployment observed in the polyamide prototypes. The trial shows a material and connector stiffness limit of the present workflow. Scale bar: \mbox{\(1~\mathrm{cm}\)}.}
    \label{fig:supp_metal_materials}
\end{figure}

An additional aluminum laser-cut trial was also tested with the same connector idea used for the polyamide prototypes. Fig.~\ref{fig:supp_metal_materials} shows the result. The sheet could be laser cut, but the connector regions were not flexible enough to allow the intended panel rotation and deployment observed with polyamide.

\subsection{Dataset generation}
\label{supp:data_generation}
Feasible target-design pairs are generated on a fixed \(10\times 10\) grid. A Sobol base-\(2\) quasi-random sequence is first drawn on \([-1,1]^{10\times 10}\)~\citep{sobol1967distribution} for more even coverage of the ratio-field space than plain random sampling. Each entry is then mapped through \(x_{ij}=10^{z_{ij}}\), which gives the ratio-field range \([1/10,10]\). Each candidate is decoded, laid out at the deployment setting used for silhouette matching, recentered by the bounding-box midpoint, and rasterized to a \(128\times 128\) binary silhouette mask. A candidate is discarded if reconstruction fails, if any invalid quadrilateral is detected, if the estimated overlap ratio exceeds \(0.02\), or if the rasterized mask fails simple topology checks. A held-out test split is generated independently, and dataset verification re-renders \(128\) random samples per split and requires IoU at least \(0.999\). The fixed \(5000/500/500\) train/validation/test split reported in Sec.~\ref{subsec:exp_setup} is then used.

\subsection{Experiment provenance and setup}
\label{supp:experiment_setup}
The reported runs use the train/validation/test split described in Sec.~\ref{subsec:exp_setup}.
Sec.~\ref{exp:fm_vs_solver} uses the full test split. The overlap threshold is \(\tau_{\mathrm{ov}}=0.02\), and the success threshold is \(\tau_{\mathrm{sIoU}}=0.85\). Each stochastic solver run uses a fixed per-target random state. Solver stopping uses the same tolerance-based settings as Table~\ref{tab:fm_vs_solver}. During each solver run, up to \(12\) feasible candidates are kept, and the one with the best final Procrustes-aligned \(\mathrm{sIoU}\) is reported. PSO starts from uniformly sampled particles with zero initial velocity. Random-restart local search uses Gaussian proposal batches with a shrinking step size. CMA-ES starts from a random mean with initial step size \(0.3\) times the box width. The nonlinear search baseline inspired by Dudte et al.~\citeyearpar{dudte2023additive} is run as a bounded Powell search with one random start~\citep{powell1964efficient}. The OT-CFM prior row in Table~\ref{tab:fm_vs_solver} uses the same Euler-\(8\) \(K=1\) setting as Table~\ref{tab:gen_models}. Fig.~\ref{fig:inverse_design_summary}(c,d) uses the full test split, with both solver baselines and OT-CFM shown to \(K=32\). Training this OT-CFM prior took about \(2~\mathrm{h}\) on the hardware in Appendix~\ref{supp:hardware}.
Fig.~\ref{fig:inverse_design_summary}(b) uses Heart, Circle, and Hexagon target masks rendered at \(128\times 128\). The solver rows use the same tolerance-based solver settings as Table~\ref{tab:fm_vs_solver}, averaged over the three masks. For each grid size, the OT-CFM prior was trained using the same recipe as the main model, with the input and output ratio-field size changed to match the selected grid. The OT-CFM row measures one Euler-\(8\) OT-CFM sample plus evaluation per target.

Sec.~\ref{exp:gen_model_choice} uses the full test split. Sec.~\ref{exp:rl_acc_phys} reports one final evaluation on the full test split, but GRPO fine-tuning rollouts use the training split and validation is used only for model selection. Fig.~\ref{fig:gen_steps_sensitivity}(a,b) uses the same trained diffusion and OT-CFM prior models as Table~\ref{tab:gen_models}, with the chosen settings from the table highlighted in the plots. The rows in Tables~\ref{tab:gen_models} and~\ref{tab:rl_results_new} correspond to the reported cGAN, diffusion, OT-CFM prior, and RL-Kirigami settings. Table~\ref{tab:rl_results_new} and Fig.~\ref{fig:table3_sensitivity}(a,b) report each RL variant at a common budget of \(10000\) environment calls.
Appendix~\ref{supp:hyperparams} summarizes the key hyperparameters used in these runs.

\subsection{Silhouette similarity metric}
\label{supp:siou_metric}
The reported \(\mathrm{sIoU}\) is computed from the simulated and target binary masks. For each candidate, the predicted mask is built by decoding \(\mathbf{x}\), laying out the deployed geometry at the evaluation setting, recentering it by the bounding-box midpoint, and rasterizing the union of quads to a \(128\times 128\) mask. The predicted and target masks are then thresholded at \(0.5\) and converted to foreground point sets. A similarity transform with translation, rotation, and isotropic scale is estimated to align the prediction to the target, because this comparison is meant to measure shape agreement and not absolute position, orientation, or overall size. The implementation uses a Procrustes-style alignment~\citep{gower1975generalized}, with a boundary-point initialization when enough boundary points are available and otherwise a principal-axis initialization, followed by one nearest-neighbor refinement step. The aligned prediction is rasterized back to the image grid, and \(\mathrm{sIoU}\) is the IoU between this aligned mask and the target mask.

For RL-Kirigami, this \(\mathrm{sIoU}\) term is combined only with feasibility penalties for reconstruction failure, invalid quadrilaterals, and overlap beyond \(\tau_{\mathrm{ov}}\). There is no separate term that directly penalizes or favors the entries of \(\mathbf{x}\). The regularity-focused runs add only the TV term from Sec.~\ref{exp:rl_acc_phys}.

\section[Additional analyses]{Additional analyses}
\label{supp:additional_experiments}

\subsection[Design resolution ablation]{Design resolution ablation}
\label{supp:design_resolution_ablation}
This ablation treats the square grid size \mbox{\(N\times N\)} as the design resolution for the finite sheet setting. Table~\ref{tab:supp_design_resolution_ablation} reports four grid sizes under the same target family and evaluation setup.

At each grid size, the OT-CFM prior was trained using the same recipe as the main model, with the input and output ratio-field size changed to match the selected grid.

\begin{table}[pos=t]
    \centering
    \caption{Design resolution ablation for finite sheet silhouette matching.
    The grid size \mbox{\(N\times N\)} controls the number of ratio-field design variables.
    Each entry uses the same target mask resolution, deployment setting, ratio bounds, decoder, and feasibility checks.}
    \label{tab:supp_design_resolution_ablation}
    \footnotesize
    \renewcommand{\arraystretch}{1.08}
    \begin{tabular}{@{}ccccc@{}}
        \toprule
        Grid & Design variables & \mbox{\(\mathrm{sIoU}\)} (\%) & \mbox{\(p_{\mathrm{succ}}\)} (\%) & Time/target (s) \\
        \midrule
        \mbox{\(6\times 6\)} & 36  & 91.7 & 92.6 & 0.14 \\
        \mbox{\(8\times 8\)} & 64  & 91.5 & 96.0 & 0.15 \\
        \mbox{\(10\times 10\)} & 100  & 94.2 & 98.7 & 0.14 \\
        \mbox{\(12\times 12\)} & 144  & 92.4 & 94.0 & 0.14 \\
        \bottomrule
    \end{tabular}
\end{table}

Among the four tested grids, the \mbox{\(10\times 10\)} setting gives the highest \mbox{\(\mathrm{sIoU}\)} and \mbox{\(p_{\mathrm{succ}}\)}, with \mbox{\(94.2\%\)} and \mbox{\(98.7\%\)}, while the per target time remains between \mbox{\(0.14\)} and \mbox{\(0.15\)}~s. This supports using \mbox{\(10\times 10\)} as the default design resolution for the reported experiments.

\subsection[OOD failure cases]{OOD failure cases}
\label{supp:ood_failure_cases}
\begin{figure}[pos=!tbp]
    \centering
    \includegraphics[width=0.5\textwidth]{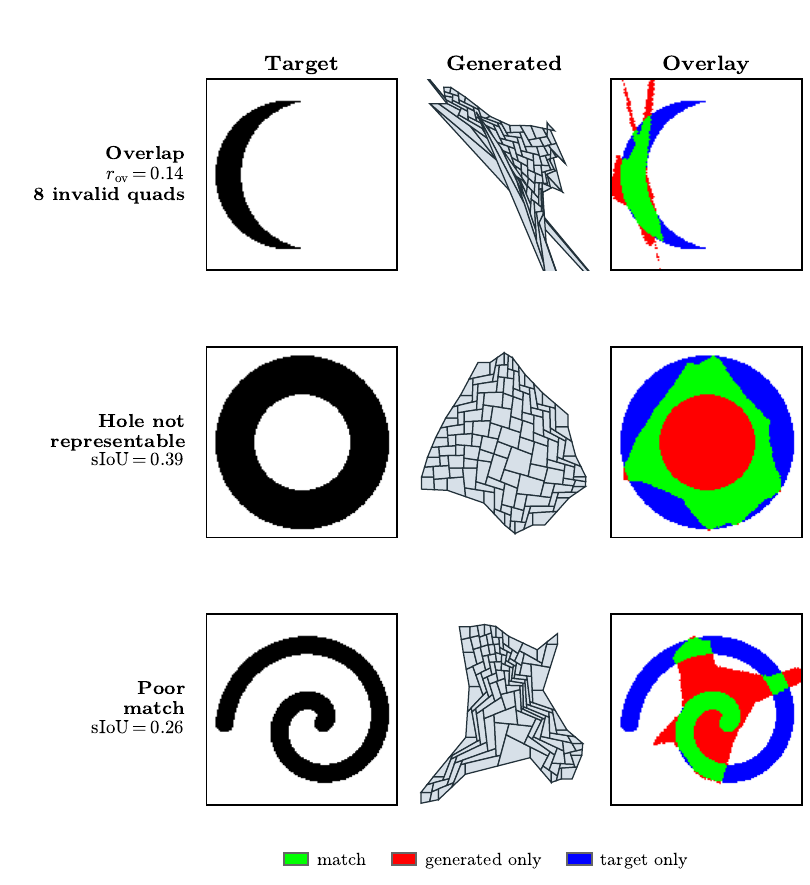}
    \caption{Representative OOD failure cases, one per mode. Each row shows the target silhouette, the generated deployed structure, and an overlay of the generated silhouette against the target, where green is matched area, red is generated only, and blue is target only. Top: overlap, where decoded quads cross and the feasibility filter rejects the design. Middle: hole mismatch, where the generated silhouette fills the target hole and the overlay shows a red disc inside a blue ring. Bottom: poor silhouette match, where a feasible design does not follow a long, thin target. The left labels report the overlap ratio \mbox{\(r_{\mathrm{ov}}\)} and invalid quadrilateral count for the overlap row, and \mbox{\(\mathrm{sIoU}\)} for the other rows. These feasibility measures are defined in Sec.~\ref{sec:param_marching}. The ratio \mbox{\(r_{\mathrm{ov}}\)} measures decoded void area lost to mutual overlap, so larger values indicate stronger self-collision.}
    \label{fig:supp_failure_cases}
\end{figure}

The OOD robustness evaluation in Sec.~\ref{exp:ood_silhouette} assigns failure categories to interpret targets with lower scores. Fig.~\ref{fig:supp_failure_cases} shows one representative case for each category.

Three failure modes are used. Overlap denotes decoded parallelogram quads that cross one another, so the flat cut pattern cannot deploy without self-collision. It is indicated by a high overlap ratio \mbox{\(r_{\mathrm{ov}}\)} and a nonzero invalid quadrilateral count, as in the thin crescent example. Hole mismatch denotes a target with an interior hole for which the generated structure fills the hole and limits \mbox{\(\mathrm{sIoU}\)}, as in the ring example. This case is interpreted as a limitation of the current run and topology, not as proof that all targets with holes are infeasible. Poor silhouette match denotes a feasible decoded structure whose boundary cannot follow a long, thin target such as the spiral. These categories help explain the lower scores of the concave and literal groups and the group with holes in Fig.~\ref{fig:ood_overview}.

\section{Tables and Figures}
\label{supp:tables_figures}

\subsection{Key experiment hyperparameters}
\label{supp:hyperparams}
This subsection collects the main settings for the final generator, RL, and solver runs reported in the study. Table~\ref{tab:supp_hyperparams} groups the shared backbone settings, method-specific training choices, RL budgets, and solver search settings in one place.

\begin{table}[pos=t]
    \centering
    \caption{Key hyperparameters for the reported experiments.}
    \label{tab:supp_hyperparams}
    \footnotesize
    \renewcommand{\arraystretch}{1.08}
    \setlength{\extrarowheight}{1pt}
    \setlength{\tabcolsep}{4pt}
    \begin{tabular}{@{}p{4.2cm}p{10.9cm}@{}}
        \toprule
        Setting & Value \\
        \midrule
        \multicolumn{2}{@{}l}{\textbf{Shared OT-CFM / diffusion backbone}} \\
        Input, mask, and latent sizes & \(10\times 10\) input, \(128\times 128\) mask, \(32\times 32\) latent grid; channels \(\{64,128,256,384\}\), \(2\) residual blocks per level, attention in the last \(3\) levels \\
        Normalization and transformer details & Group norm with \(32\) groups, head widths \(\{32,64,128,192\}\), \(12\) transformer layers, dropout \(=0.1\), maximum timestep \(=1000\) \\
        \midrule
        \multicolumn{2}{@{}l}{\textbf{Generator-specific settings}} \\
        OT-CFM training and sampling & AdamW, learning rate \(2\times 10^{-5}\), weight decay \(0.05\), batch \(64\), \(400\) epochs; OT coupling; Euler sampling with \(9\) time points and step size \(1/8\); SWA learning rate \(2\times 10^{-6}\) \\
        Diffusion training and sampling & Scaled linear beta schedule with \((5\times 10^{-4},\,1.95\times 10^{-2})\), DDIM with \(25\) inference steps \\
        cGAN architecture and training & U-Net generator with channels \(\{32,64\}\), stride \(2\), \(1\) residual unit; patch discriminator with \(32\) channels and \(1\) layer; Adam with \((0.5,0.999)\), \(\lambda_{L1}=10\), batch \(64\), \(400\) epochs \\
        cVAE architecture and training & \(32\times 32\) conditional input, \(10\times 10\) output, \(4\) latent channels, KL weight \(10^{-4}\), L1 reconstruction, Adam \(2\times 10^{-4}\), batch \(64\), \(400\) epochs \\
        \midrule
        \multicolumn{2}{@{}l}{\textbf{GRPO fine-tuning}} \\
        Shared settings & OT-CFM prior initialization, learning rate \(10^{-5}\), batch \(8\), group size \(4\), reward temperature \(0.2\), overlap threshold \(0.02\) \\
        Selected runs & Accuracy, regularity-only, and accuracy plus regularity are all reported at a common budget of \(10000\) environment calls. The accuracy run uses no TV penalty, the regularity-only run uses a TV-only reward, and the accuracy plus regularity run uses equal normalized \(\mathrm{sIoU}\) and TV contributions \\
        \midrule
        \multicolumn{2}{@{}l}{\textbf{Solver baselines}} \\
        Common settings & \(10\times 10\) bounded search field with box bounds \([0.1,10.0]\), \(x\)-tolerance \(10^{-3}\), relative objective tolerance \(10^{-3}\), patience \(5\), \(1000\)-evaluation safety cap, \(\tau_{\mathrm{sIoU}}=0.85\), \(\tau_{\mathrm{ov}}=0.02\) \\
        Search settings & PSO swarm size \(24\) with inertia \(0.7\) and \(c_1=c_2=1.5\); random-restart local search with \(8\) restarts, batch \(10\), initial step \(0.25\), and shrink factor \(0.5\); CMA-ES population \(4+\lfloor 3\log d\rfloor\) with initial step size \(0.3\) of the box width; Powell-style search with one random start under the same tolerance-based stop \\
        \bottomrule
    \end{tabular}
\end{table}

\section{Model Details}
\label{supp:model_details}

\subsection{Shared U-Net backbone}
\label{supp:shared_unet}
\begin{figure}[pos=t]
    \centering
    \includegraphics[width=\textwidth]{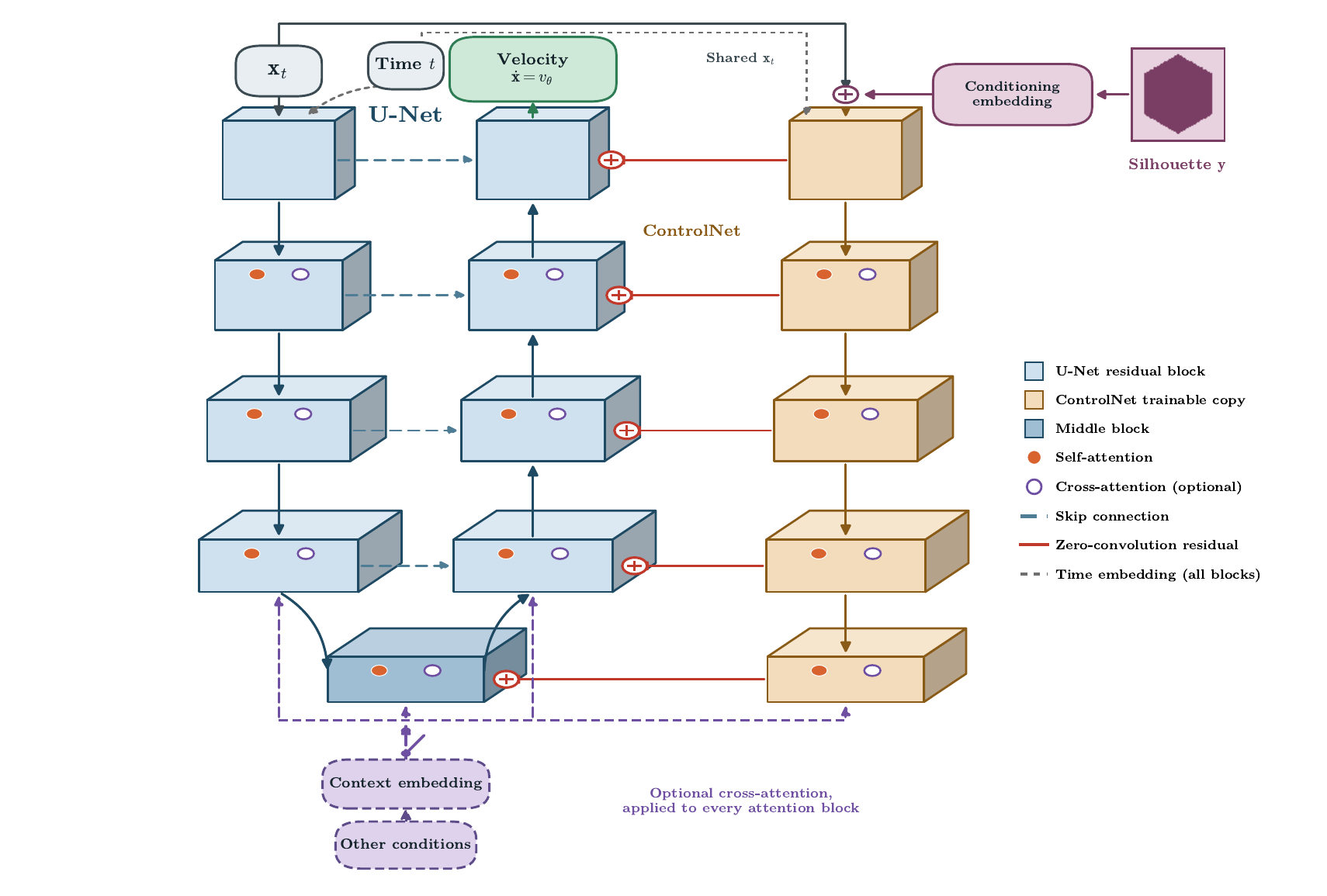}
    \caption{Shared mask conditioned U-Net backbone with optional ControlNet conditioning, drawn for the OT-CFM velocity field \mbox{\(v_\theta(\mathbf{x},t,\mathbf{y},\varphi,\mathcal{B})\)}. The latent ratio field \mbox{\(\mathbf{x}_t\)} and the time \mbox{\(t\)} enter the U-Net, which has a downsampling arm, a middle block, and an upsampling arm joined by skip connections. Self-attention is used in the last three levels and the middle block. The ControlNet is a trainable copy of the downsampling arm and middle block. It takes the same \mbox{\(\mathbf{x}_t\)} together with the embedded target silhouette \mbox{\(\mathbf{y}\)} (shown here as a hexagon), and adds zero convolution residuals to the U-Net skip samples and middle block. The time embedding is added to every residual block of both networks.}
    \label{fig:backbone}
\end{figure}

All OT-CFM and diffusion models use the same mask-conditioned diffusion U-Net backbone with optional ControlNet conditioning~\citep{zhang2023controlnet}. The U-Net is the neural network used during sampling. It is not the geometric decoder \(\mathcal{G}\). After the network predicts a field, \(\mathcal{G}\) is the separate marching decoder from Sec.~\ref{sec:param_marching} that converts the field into a kirigami layout and applies the feasibility checks. The model operates on a \(10\times 10\) field representation and a \(128\times 128\) target mask. Inside the network, the current \(10\times 10\) field is resized to a \(32\times 32\) latent grid, processed by the shared U-Net, and resized back to a \(10\times 10\) output field. When ControlNet is enabled, the target mask is also resized to \(32\times 32\) and injected through residual features at the down blocks and mid-block. Fig.~\ref{fig:backbone} shows this backbone and its conditioning paths schematically.

The shared backbone uses four resolution levels. The channel widths are \(\{64,128,256,384\}\), with two residual blocks per level and attention in the last three levels. The backbone uses group normalization with \(32\) groups, residual up and down sampling, head widths \(\{32,64,128,192\}\), \(12\) transformer layers, attention dropout \(=0.1\), and maximum timestep \(=1000\). OT-CFM uses this backbone with OT coupling, meaning optimal transport pairing between base samples and training designs during flow matching training. The final OT-CFM run uses Euler sampling with \(9\) time points and step size \(1/8\), learning rate \(2\times 10^{-5}\), weight decay \(0.05\), stochastic weight averaging, batch size \(64\), and \(400\) training epochs. The diffusion baseline uses the same shared backbone with \(1000\) training timesteps, Denoising Diffusion Implicit Models (DDIM) sampling at test time~\citep{song2021ddim}, batch size \(64\), and \(400\) training epochs.

\subsection{Other model details}
\label{supp:other_model_details}

The cGAN generator in the final comparison uses a conditioned convolutional generator with channel widths \(\{32,64\}\), stride \(=2\), one residual unit, PReLU activations, and instance normalization. The cGAN discriminator is a one-layer patch discriminator with \(32\) channels, and the final cGAN run uses batch size \(64\) and \(400\) training epochs. A conditional VAE baseline built from a SPADE autoencoder KL~\citep{park2019semantic} is also included. It uses a \(32\times 32\) conditional input grid, a \(10\times 10\) output grid, latent channels \(=4\), KL weight \(10^{-4}\), L1 reconstruction loss, prior sampling at test time, batch size \(64\), and \(400\) training epochs.

For RL-Kirigami, the starting point is the OT-CFM prior used in Sec.~\ref{exp:gen_model_choice}. Euler-\(8\) sampling is kept during rollout and evaluation. The three RL variants reported in Table~\ref{tab:rl_results_new} and Fig.~\ref{fig:table3_sensitivity}(a,b) use group size \(4\), batch size \(8\), and a common budget of \(10000\) environment calls. The accuracy row uses \(\mathrm{sIoU}\) reward with no TV penalty. The regularity-only row uses a TV-only reward, while the accuracy plus regularity row uses equal normalized \(\mathrm{sIoU}\) and TV contributions, so half of the hybrid reward comes from each term. In all runs, reconstruction failure uses penalty \(5.0\), invalid-quadrilateral presence uses penalty \(2.0\), overlap beyond \(\tau_{\mathrm{ov}}\) uses weight \(2.0\), and there is no separate term on the entries of \(\mathbf{x}\).

\end{document}